%% file: main.tex
\documentclass[10pt]{cyclegrponus}
\usepackage{float,threeparttable,adjustbox,colortbl,makecell}
\usepackage[linesnumbered,lined,boxed,commentsnumbered,ruled]{algorithm2e}
\usepackage[misc]{ifsym}
\usepackage{pifont,xspace}
\renewcommand{\emph}[1]{\textit{#1}}
\providecommand{\IEEEPARstart}[2]{#1#2}
\input{paper_macros}
\colorlet{blue}{nusblue}
\hypersetup{pdftitle={Generalized CycleGRPO for Fine-Grained Multimodal Understanding and Grounding},
 pdfauthor={Xin Zhang, Haochen Wang, Beibei Lin, Zhaoxiang Zhang, Robby T. Tan, Dacheng Tao, Jason Li},
 linkcolor=nusblue,citecolor=nusblue,urlcolor=nusblue}
\renewcommand{\authorfont}{\fontsize{12}{14}\selectfont}
\renewcommand{\affiliationfont}{\fontsize{9}{12}\selectfont}
\title{Generalized CycleGRPO for Fine-Grained\\Multimodal Understanding and Grounding}
\author[]{CycleGRPO Team: NUS, UCAS, NTU}
\abstract{\input{secs/0_abstract}}
\begin{document}
\maketitle
\suppressfloats[t]
\input{secs/0_teaser}

\input{secs/1_introduction}
\input{secs/2_related_work}
\input{secs/3_method}
\input{secs/4_benchmark}
\input{secs/4_exp}
\input{secs/5_limitations}
\input{secs/5_conclusion}
\input{secs/contributors}
\clearpage
\begingroup\small

\input{references}
\endgroup
\clearpage
\beginappendix
\input{secs/appendix_overview}

\input{secs/6_appendix_exp_settings}
\end{document}

%% file: paper_macros.tex
\newcommand*{\belowrulesepcolor}[1]{
  \noalign{
    \kern-\belowrulesep 
    \begingroup 
      \color{#1}
      \hrule height\belowrulesep 
    \endgroup 
    
    \vspace{-0.03mm}
  }
} 
\newcommand*{\aboverulesepcolor}[1]{
  \noalign{
  \vspace{-0.03mm}
    \begingroup 
      \color{#1}
      \hrule height\aboverulesep 
    \endgroup 
    \kern-\aboverulesep 
  }
}

\usepackage{pifont}
\newcommand{\cmark}{\ding{51}}
\newcommand{\xmark}{\ding{55}}

\usepackage{xspace}
\makeatletter
\DeclareRobustCommand\onedot{\futurelet\@let@token\@onedot}
\def\@onedot{\ifx\@let@token.\else.\null\fi\xspace}

\makeatother

%% file: secs/0_abstract.tex
Fine-grained multimodal intelligence requires a model not only to understand a designated region, but also to localize that region from language. Existing methods learn these two directions with separate objectives and task-specific annotations. We present \emph{Generalized CycleGRPO}, the first unified reinforcement learning framework that couples region understanding and localization across spatial, temporal, and spatio-temporal domains. Given a region, the multimodal large language model first generates candidate descriptions and then localizes each description to reconstruct the input region. The overlap between the input and reconstructed regions provides a geometrically verifiable reward for description quality, while rewards for individual localization trajectories improve localization. We represent regions in a common space equipped with an overlap measure. This allows the same learning principle to apply to image masks or boxes, video intervals, and timestamped spatial evidence, without caption annotations or an external language-model judge during reinforcement learning. For videos, this cycle further rewards descriptions that distinguish a target event from semantically similar recurrences. Accordingly, we introduce DOVE (Distinctive Occurrence-level Video-caption Evaluation), which evaluates this fine-grained distinctiveness through contrastive groups of target intervals and same-query recurrences. Across image benchmarks, CycleGRPO raises SAMTok's overall gIoU from 57.5\% to 68.2\% on GroundingSuite and its average gIoU from 71.2\% to 78.7\% on GRES. In the temporal setting, it reaches 51.9\% tIoU on Charades and 33.3\% Temporal Evidence Coverage (TEC) on DOVE, compared with 45.6\% and 14.8\% for its Qwen3-VL initialization. In the joint spatio-temporal setting, it raises V-STaR mAM from 34.6\% to 36.2\% and mLGM from 47.2\% to 49.8\%. These results establish region--text cycle consistency as a general self-evaluation principle for jointly advancing multimodal understanding and localization. Models, code, and the DOVE benchmark are available at \url{https://devinxzhang.github.io/Generalized-CycleGRPO-Page/}.

%% file: secs/0_teaser.tex
\begin{figure*}[t]
    \centering
    \includegraphics[width=\textwidth]{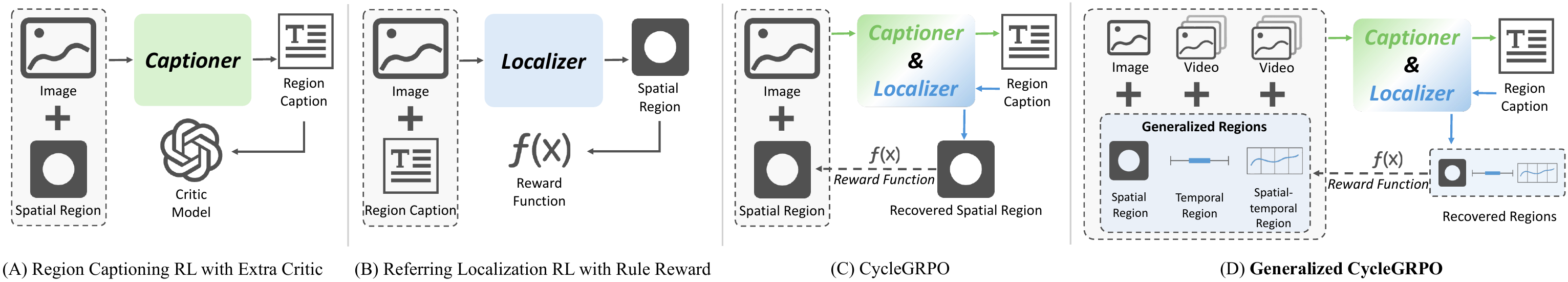}
    \caption{From spatial CycleGRPO to Generalized CycleGRPO.
    (A) Region-captioning reinforcement learning relies on an extra critic to
    score generated descriptions. (B) Referring-localization reinforcement
    learning receives a rule-based geometric reward but optimizes only the
    localization direction. (C) Our spatial CycleGRPO closes the
    region--caption--localization loop and uses the overlap of the recovered
    spatial region as a verifiable reward. (D) This work generalizes the same
    cycle over spatial, temporal, and spatio-temporal regions, unifying
    captioning and localization for images and videos without an external
    critic.}
    \label{fig:teaser}
\end{figure*}

%% file: secs/1_introduction.tex
\section{Introduction}
\label{sec:intro}

\IEEEPARstart{L}{arge} language models are being explored for scientific discovery and hypothesis generation~\cite{chen2025auto,chen2025hypospace}, while multimodal large language models (MLLMs) have rapidly
progressed from holistic image description to open-ended visual reasoning and
video understanding~\cite{llava,qwen25vl,Qwen3-VL,team2026kimi,liu2026tele}. Yet many
questions are not about an entire image or video: they concern a particular
object, spatial region, temporal event, or tracked entity across space and time~\cite{zhang2025mamba,zhang2023adaptive,zhang2024heap,yang2025erf,Li_2026_CVPR,li2026bridging,DA-LLPose,chen2026udapose,cao20263dot,yan2025erlora,Yan_2025_CVPR,Yan_2026_CVPR,zhang20264dpc2hatdynamicpointcloud,tan2024rle,tan2026knowing,tan2024occluded,yuan2026drive,yuan2025tailored,qian2024msnsegnet,zhang2022gigapixel,lin2025rgb,lin2026polarscale}.
Answering them requires two complementary capabilities. 
A model must \emph{understand} a designated region well enough to describe its content, and must also \emph{localize} that content by mapping language back to a spatial or temporal
extent. 
As illustrated in Fig.~\ref{fig:teaser}(A)--(B), existing reinforcement
learning formulations typically optimize these two directions separately.
Although recent MLLMs support masks, boxes, and timestamps as inputs or outputs~\cite{glamm,zhou2026samtok,videorefer,zhang2026timelens,sa2va,ren2024pixellm}, this separation remains common.

This separation creates an avoidable supervision bottleneck. Region understanding is commonly trained against expensive textual annotations, whereas localization is trained against geometric targets. 
The two forms of supervision describe the same underlying correspondence but do not constrain one another: a fluent caption need not uniquely identify its source region, and a strong localizer provides no direct signal to improve the captions it consumes.  
This is particularly limiting at fine granularity, where distinguishing objects, actions, and their interactions requires more than coarse category recognition~\cite{lei2025hola,lei2024ezhoi}. A useful description must state the details that distinguish one occurrence from its confounders, not merely remain plausible in isolation.

Our ECCV paper introduced CycleGRPO for image regions and
showed that this duality can replace textual supervision during reinforcement
learning~\cite{zhang2026cyclegrpo}.
As shown in Fig.~\ref{fig:teaser}(C), spatial CycleGRPO couples the two
directions through a closed region--caption--localization loop.
Starting from a mask or box, the model generates a caption and subsequently
uses that caption to localize the region in the image.
Reconstruction IoU serves as a verifiable reward: descriptions are preferred when they recover the original region, while the localization rollouts are themselves optimized against the known region. 
This spatial formulation substantially improves both region understanding and referring localization. 
This leaves an open question: can the same cycle extend beyond two-dimensional images to other region spaces?

In this work, we address this question with \emph{Generalized CycleGRPO}, a geometry-agnostic framework spanning \emph{spatial}, \emph{temporal}, and \emph{spatio-temporal} region spaces.
As illustrated in Fig.~\ref{fig:teaser}(D), we represent a region abstractly as $\mathcal{R}\in\Omega$ and optimize the cycle $\mathcal{R}\rightarrow\mathcal{C}\rightarrow\widehat{\mathcal{R}}$ using an overlap operator defined over $\Omega$.
Defining $\Omega$ as a space of image masks or bounding boxes recovers spatial CycleGRPO. Temporal intervals yield temporal captioning and localization, while timestamped boxes, masks, or region tubes support joint when-and-where localization.
The same policy plays both roles: it acts on a provided region to sample candidate descriptions, then acts as its own critic by localizing those descriptions. 
Group-relative optimization ranks the descriptions by reconstruction quality and trains each localization trajectory from its geometric overlap. 
Consequently, the reinforcement learning stage requires region annotations alone---no reference captions, external reward model, or external LLM judge.

Extending the cycle from images to videos is not a mechanical replacement of
spatial IoU with temporal IoU. 
First, an event may recur several times in a video, so a generic but factually correct caption can ground to multiple intervals and fail to define a one-to-one correspondence.  This makes \emph{temporal distinctiveness}---stating the context unique to the target
occurrence---central to cycle consistency. 
Second, visual evidence has two coupled coordinates: the model must identify \emph{when} an event occurs and \emph{where} its supporting objects appear. 
Open-o3-Video~\cite{meng2025openo3video}, for example, learns from spatio-temporal reasoning traces and produces timestamps and boxes as explicit evidence.
For the first challenge, the temporal instantiation rewards a description only
when it reconstructs the designated interval rather than a similar recurrence,
thereby favoring occurrence-specific state, order, and outcome cues.  
For the second, the spatio-temporal instantiation represents a region as timestamped
spatial evidence and evaluates spatial overlap only at aligned timestamps.
Thus, neither a similar object at the wrong time nor the correct time without
the relevant spatial evidence can satisfy the cycle.

These video settings also expose an evaluation gap. 
Conventional captioning metrics reward similarity to a single reference, while temporal-grounding benchmarks evaluate only the inverse direction. 
Neither determines whether a caption identifies its own occurrence rather than a semantically similar segment elsewhere in the same video. We therefore introduce \emph{DOVE} (Distinctive Occurrence-level Video-caption Evaluation), constructed from videos containing multiple occurrences of a shared query.
Its contrastive protocol measures whether a caption is faithful to its target and whether its specific facts exclude neighboring occurrences. 
We summarize its coverage of order, state, and outcome evidence using \emph{Temporal Evidence Coverage} (TEC), which macro-averages coverage over these three probe types.
DOVE complements standard grounding evaluation and directly tests the one-to-one region--text correspondence.

\begin{figure}[!t]
    \centering
    \includegraphics[width=0.9\columnwidth]{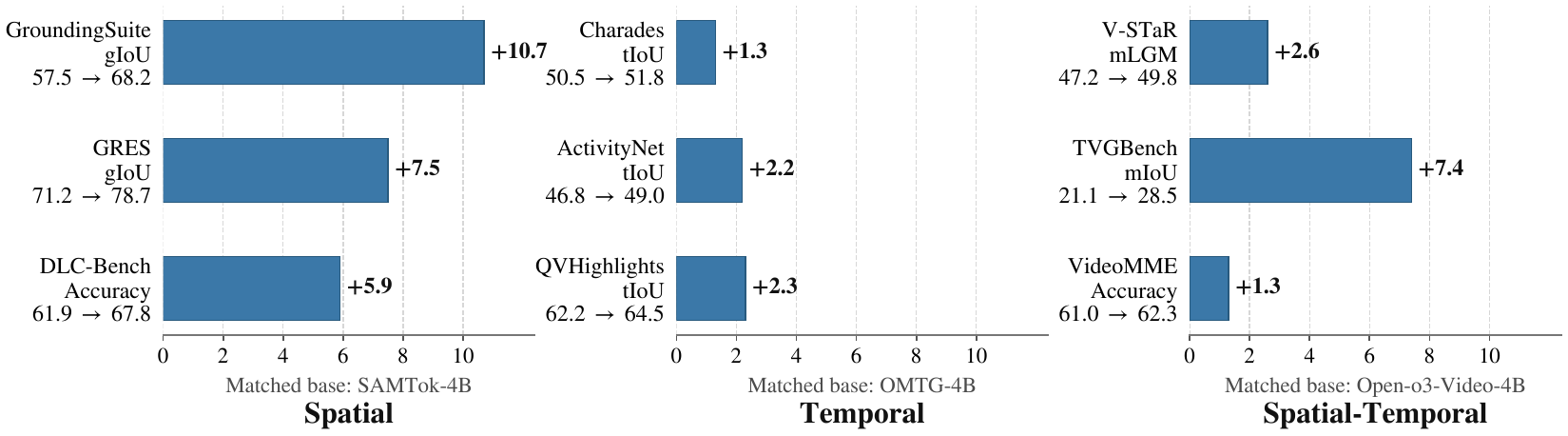}
    \caption{Improvements over the corresponding base models on spatial
    (left), temporal (middle), and spatio-temporal (right) benchmarks.
    Within each benchmark, bar lengths show gains in percentage points and
    labels give the base and CycleGRPO scores.}
    \label{fig:matched_gains}
\end{figure}

We evaluate the framework across all three region spaces.
In the spatial setting, CycleGRPO raises SAMTok's overall gIoU from
57.5\% to 68.2\% on GroundingSuite and its average gIoU from 71.2\% to 78.7\%
on GRES; DLC-Bench accuracy also improves from 61.9\% to 67.8\%. 
Starting from the task-specialized OMTG-4B temporal model, Generalized CycleGRPO raises
tIoU from 50.5\% to 51.8\% on Charades, from 46.8\% to 49.0\% on ActivityNet,
and from 62.2\% to 64.5\% on QVHighlights. 
In the joint spatio-temporal
setting, it improves Open-o3-Video-4B's V-STaR mLGM from 47.2\% to 49.8\% and
TVGBench mIoU from 21.1\% to 28.5\%, while also raising VideoMME accuracy from
61.0\% to 62.3\%. 
Together, these results show that localization is not merely a critic for language generation:
both directions can be bootstrapped within one verifiable cycle, with
domain-dependent gains and without reference captions in the cycle branches
during reinforcement learning.
Fig.~\ref{fig:matched_gains} summarizes these gains against the
task-specific matched base in each region space.

This paper extends our ECCV version in four ways:
\begin{itemize}
    \item We extend CycleGRPO beyond the spatial masks and boxes studied in the ECCV
    version to a generalized region--text cycle over spatial masks/boxes,
    temporal intervals, and spatio-temporal trajectories, unified through
    region-specific serializers and consistency functions.
    \item We introduce temporal and spatio-temporal training instantiations in
    which one shared policy learns segment description, interval localization,
    and temporally aligned spatial evidence from region annotations alone,
    without reference captions or an external judge during reinforcement
    learning.
    \item We introduce DOVE, a human-verified contrastive benchmark for
    distinctive occurrence-level video-caption evaluation. 
    For the first time, it measures whether segment captions distinguish their target from same-query recurrences.
    \item We provide comprehensive validation across the three region spaces,
    including single-task and reward-side ablations, backbone comparisons,
    and analyses of temporal distinctiveness and failure modes across more than ten benchmarks.
\end{itemize}

%% file: secs/2_related_work.tex
\section{Related Work}
\label{sec:related_work}

\noindent
\textbf{Image Region Understanding and Grounding.}
Fine-grained image understanding and localization involve two inverse mappings. Region
understanding conditions a description or answer on a designated
region~\cite{chen2023shikra,peng2023kosmos,you2023Ferret,ferret2,osprey,guo2024regiongpt},
as in GLaMM~\cite{glamm}, DAM~\cite{dam}, and GAR~\cite{wang2025grasp}.
Referring localization maps free-form language to a box or
mask~\cite{gres,lisa,ren2024pixellm,du2026weatherreasonseg}, via unified architectures such as
OMG-LLaVA~\cite{zhang2024omg} and Sa2VA~\cite{sa2va} or discrete mask
sequences~\cite{himtok,zhou2026samtok}. Our implementation builds on
SAMTok~\cite{zhou2026samtok}, which represents a mask with two region tokens so
that one MLLM can both consume and emit spatial regions.
Supporting both directions in one architecture does not couple their learning
signals: descriptions and grounded masks are optimized with different
objectives or annotation sources. 
Our conference work~\cite{zhang2026cyclegrpo} connects them through the cycle region$\rightarrow$text$\rightarrow$region; the present work treats this as one
instance of a geometry-independent principle and extends it to temporal and spatio-temporal regions.

\noindent
\textbf{Temporal and Spatio-Temporal Understanding and Grounding.}
Early video MLLMs emphasize holistic description and question
answering~\cite{maaz2024videochatgpt,zhang2023videollama,xu2024pllava,li2024llavaonevision},
with recent backbones supporting long videos and timestamp
conditioning~\cite{qwen25vl}. At temporal granularity, language-conditioned
grounding localizes an event interval~\cite{huang2024vtimellm,ren2024timechat,qian2024momentor,huang2024lita,guo2024vtgllm,guo2025trace},
while the inverse task describes a designated
interval~\cite{wang2021pdvc,yang2023vid2seq,videorefer}. 
These approaches demonstrate useful interaction between localization and captioning, but learn
it from paired timestamp--caption supervision in a forward prediction objective rather than from a closed, self-reconstructive region--text cycle.
Spatio-temporal grounding further asks \emph{when} a referred entity occurs and
\emph{where} it appears. 
Prior methods~\cite{zhang2020vidstg,yang2022tubedetr,videorefer,sa2va} map language to tubes, decode frame-level boxes, or predict mask sequences.
Moving to evidence-centered reasoning, Open-o3-Video emits timestamps and boxes
alongside its answer and optimizes answer, temporal, and spatial
rewards~\cite{meng2025openo3video}. 
These works predominantly optimize the language$\rightarrow$region direction using paired evidence or reasoning annotations. 
Our framework additionally learns the inverse region$\rightarrow$language direction and couples the two through reconstruction in the same region space. 
In videos, repeated occurrences make this coupling nontrivial: a generic description may match several intervals, making temporal distinctiveness essential for faithful reconstruction.

\noindent
\textbf{Verifiable Reinforcement Learning and Cycle Consistency.}
Outcome-level feedback trains language models through preference and
group-relative policy optimization~\cite{rlhf,dpo,grpo}. Multimodal extensions use several forms of feedback: IoU-style rewards for detection, referring, and
segmentation~\cite{visualrft,vlmr1,segzero,segr1,univgr1}; caption rewards from
a separate vision-free answerer~\cite{caprl,yang2026caprl++}; and video
self-evolution from temporal perturbation, discovered evidence, or internal
confidence~\cite{huang2026evovid,zhang2026video,intuitor2025}. 
None, however, uses language-to-region reconstruction by the same policy to train the inverse
region-to-language mapping.
Reducing reliance on paired targets has also been explored through
teacher-based video restoration~\cite{chen2024dualrain} and cycle
consistency~\cite{zhu2017cyclegan,sennrich2016backtranslation}. CycleGRPO
instantiates the latter within one multimodal policy: the model describes a region,
then localizes that description back as its own critic. 
Reconstruction overlap rewards understanding, and each localization rollout is rewarded against the
known region, optimizing both directions without reference captions or an external judge during RL.

\noindent
\textbf{Benchmarking Fine-Grained Region--Text Correspondence.}
Existing video benchmarks evaluate either holistic understanding~\cite{li2024mvbench,fu2024videomme,liu2024tempcompass,liu2024etbench}
or one direction of fine-grained correspondence: temporal-grounding
benchmarks~\cite{gao2017charades,krishna2017activitynet,lei2021qvhighlights,grauman2022ego4d},
including TimeLens~\cite{zhang2026timelens}, score localization alone, while
captioning is scored by similarity to references. 
DLC-Bench~\cite{dam} moves beyond reference matching by checking atomic facts in region captions, while CrossHOI-Bench~\cite{lei2026crosshoibench} uses curated multiple-choice questions to evaluate human--object interaction recognition across VLMs and specialized methods.
None of these protocols tests what a temporal region--text cycle requires:
whether a caption is faithful to its own interval and discriminative against similar recurrences in the same video. 
Our DOVE benchmark fills this gap with human-verified repeated-occurrence
groups and private Order, State, and Outcome probes. 
A fixed text-only judge determines whether each target-conditioned caption supports the correct answer, supports a wrong answer, or provides insufficient evidence. This yields a probe-level assessment without reference matching or holistic grading.

%% file: secs/3_method.tex
\section{Generalized CycleGRPO}
\label{sec:method}

\begin{figure*}[t]
    \centering
    \includegraphics[width=\textwidth]{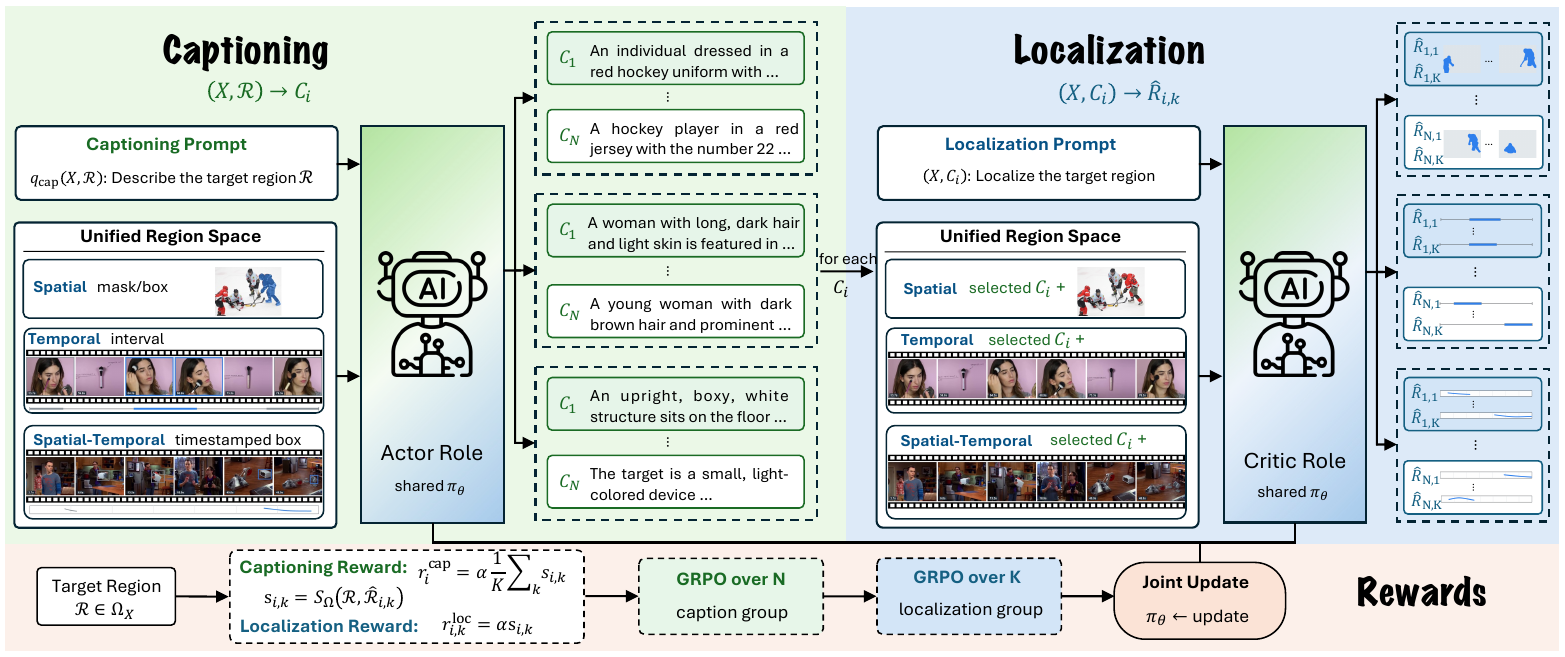}
    \caption{Overview of Generalized CycleGRPO. Given a visual input $X$ and
    a target region $\mathcal{R}$ in a unified region space, the captioning
    rollout samples $N$ candidate descriptions $C_i$. Each candidate then
    conditions $K$ localization rollouts that reconstruct regions
    $\widehat{\mathcal{R}}_{i,k}$; their agreement with the target provides
    the cycle-consistency signal. 
    Spatial masks or boxes, temporal intervals, and spatio-temporal timestamped boxes differ only in their region serialization and consistency function, allowing the same
    region--text--region optimization to operate across image and video tasks.}
    \label{fig:main_overview}
\end{figure*}

We first formulate region captioning and localization over a generic region
space, and then present the nested rollout and joint optimization used by
Generalized CycleGRPO. 
Our formulation separates the learning principle from the representation of a region: masks, boxes, temporal intervals, and spatio-temporal tubes may use different serializations and consistency
functions, but they enter the same region--text--region cycle, as summarized in
Fig.~\ref{fig:main_overview}.

\subsection{Region--Text Duality over Measurable Spaces}
Let $X$ denote a visual input, either an image or a video, and let
$\Omega_X$ denote its admissible region space. 
A target region is written as $\mathcal{R}\in\Omega_X$. Depending on the task, $\mathcal{R}$ may be a two-dimensional mask or box, a subset of the video timeline, or spatial support that varies over time. We assume that $\Omega_X$ provides (i) a
serialization through which a multimodal large language model can consume and
emit a region, and (ii) a bounded consistency function
\begin{equation}
    S_{\Omega}(\mathcal{R},\widehat{\mathcal{R}})\in[0,1],
    \qquad
    S_{\Omega}(\mathcal{R},\mathcal{R})=1,
    \label{eq:region_score}
\end{equation}
which measures how well a predicted region $\widehat{\mathcal{R}}$ recovers
$\mathcal{R}$. 
This score is a geometric overlap whenever the representation
supports direct decoding; compact region tokens may instead use a
token-consistency surrogate during training.

Region captioning and localization are inverse conditional mappings over this
space. Given $X$ and $\mathcal{R}$, region understanding generates a natural
language description $C$; given $X$ and $C$, localization reconstructs a region
$\widehat{\mathcal{R}}$:
\begin{equation}
    (X,\mathcal{R}) \xrightarrow{\;f_\theta\;} C,
    \qquad
    (X,C) \xrightarrow{\;g_\theta\;} \widehat{\mathcal{R}}.
    \label{eq:dual_mapping}
\end{equation}
Both mappings are implemented by the same autoregressive policy
$\pi_\theta$, with task-specific instruction templates. 
Region outputs are decoded by the serializer associated with $\Omega_X$; no task-specific critic
or reward model is introduced. 
The central premise is that a useful region description should be sufficiently faithful and discriminative to recover its source:
\begin{equation}
    \mathcal{R}\rightarrow C\rightarrow\widehat{\mathcal{R}},
    \qquad
    \widehat{\mathcal{R}}\simeq\mathcal{R}.
    \label{eq:generic_cycle}
\end{equation}
Our conference formulation instantiates this premise for image
regions~\cite{zhang2026cyclegrpo}. 
Generalized CycleGRPO makes the region space explicit and applies the same learning rule wherever Eq.~\ref{eq:region_score} applies.

The claim of annotation efficiency applies to the reinforcement
learning stage. 
Starting from a multimodal policy with the two required output interfaces, each CycleGRPO example contains only $(X,\mathcal{R})$; it does not contain a reference description. 
The target region therefore supervises both directions through the cycle, without a ground-truth caption, an external LLM judge, or a learned reward model.

\subsection{Nested Region--Text--Region Rollouts}
For a training pair $(X,\mathcal{R})$, we construct a region-understanding
prompt $q_{\mathrm{cap}}(X,\mathcal{R})$ and sample $N$ candidate
descriptions from the rollout policy:
\begin{equation}
    C_i\sim\pi_{\theta_{\mathrm{old}}}
    \bigl(\cdot\mid q_{\mathrm{cap}}(X,\mathcal{R})\bigr),
    \qquad i=1,\ldots,N.
    \label{eq:caption_rollout}
\end{equation}
These candidates are semantic hypotheses about the same target region. They
need not resemble a particular reference sentence; their quality is determined
by whether the policy can use them to recover the target.

Each candidate then conditions a group of inner localization rollouts.
We build $q_{\mathrm{loc}}(X,C_i)$ and independently sample $K$ region predictions:
\begin{equation}
    \widehat{\mathcal{R}}_{i,k}
    \sim\pi_{\theta_{\mathrm{old}}}
    \bigl(\cdot\mid q_{\mathrm{loc}}(X,C_i)\bigr),
    \qquad k=1,\ldots,K.
    \label{eq:localization_rollout}
\end{equation}
The model thus acts first as a description policy and then as its own
critic through localization. For every inner trajectory, we compute
\begin{equation}
    s_{i,k}=S_{\Omega}
    \bigl(\mathcal{R},\widehat{\mathcal{R}}_{i,k}\bigr),
    \qquad
    \bar{s}_i=\frac{1}{K}\sum_{k=1}^{K}s_{i,k}.
    \label{eq:cycle_score}
\end{equation}
Multiple inner rollouts are important because a single autoregressive decoding
may select one of several plausible modes. 
Averaging over these rollouts estimates how reliably $C_i$ identifies the source region.

This construction uses region reconstruction to evaluate generated descriptions.
A hallucinated description tends to reconstruct an unrelated region, while an overly generic description can be compatible with multiple objects or moments and therefore cannot reliably return to the designated one. 
This latter case is especially common in videos containing repeated instances of the same action. Maximizing $\bar{s}_i$ consequently places implicit pressure on descriptions to mention the context that distinguishes the target occurrence. 
To prevent direct coordinate copying from becoming a shortcut, we impose representation-specific anti-leakage constraints. 
In the temporal instantiation, for example, we penalize explicit timestamps in the caption and remove them before constructing the localization prompt.

\begin{figure*}[!t]
    \centering
    \includegraphics[width=\textwidth]{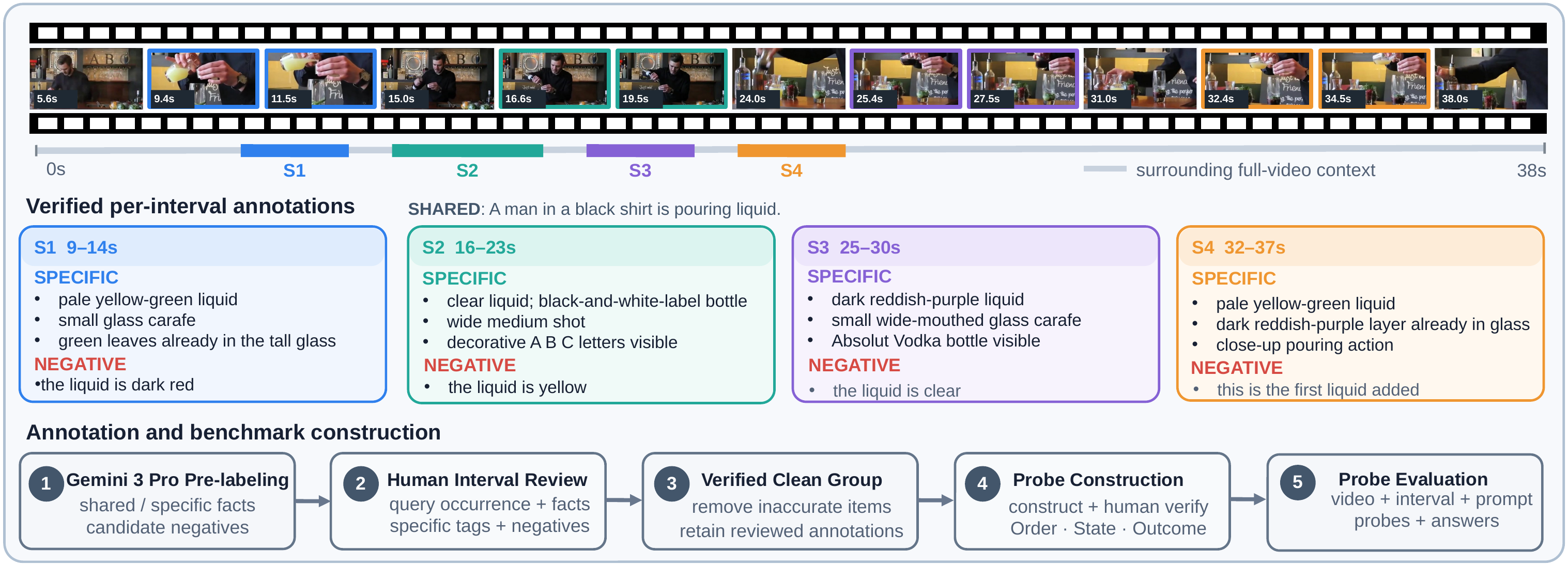}
    \caption{Construction and evaluation of DOVE. Candidate
    intervals are pre-labeled, human-reviewed, and converted into private
    temporal probes. A text-only judge then evaluates captions for specified
    intervals using Order, State, and Outcome probes.}
    \label{fig:dove-construction}
\end{figure*}

\subsection{Cycle Rewards and Joint GRPO Optimization}
The same consistency measurements provide rewards for both stages. We
define the caption reward as
\begin{equation}
    r_i^{\mathrm{cap}}
    =\alpha\bar{s}_i+\Phi_{\mathrm{cap}}(C_i),
    \label{eq:caption_reward}
\end{equation}
where $\alpha$ controls the scale of the cycle signal and
$\Phi_{\mathrm{cap}}$ contains only reference-free validity terms, such as
anti-leakage and repetition checks. Each localization trajectory receives
\begin{equation}
    r_{i,k}^{\mathrm{loc}}
    =\alpha s_{i,k}
    +\Phi_{\mathrm{loc}}(\widehat{\mathcal{R}}_{i,k}),
    \label{eq:localization_reward}
\end{equation}
where $\Phi_{\mathrm{loc}}$ checks whether the predicted region follows the
required serialization. Localization advantages are normalized within the
$K$ rollouts conditioned on a fixed parent caption, so these samples compete
on their own consistency with the target under the same linguistic description.
Cross-caption credit is supplied by Eq.~\ref{eq:caption_reward}. 
In the temporal implementation, an additional multiplicative gate downweights the
overlap term when the number of predicted intervals is invalid. 
None of these auxiliary terms requires reference text.

We use group-relative advantages separately for the two nested levels. The
$N$ descriptions of one target form a caption group, giving
\begin{equation}
    A_i^{\mathrm{cap}}
    =\frac{r_i^{\mathrm{cap}}-\mu_{\mathrm{cap}}}
    {\sigma_{\mathrm{cap}}+\delta},
    \label{eq:caption_advantage}
\end{equation}
where $\mu_{\mathrm{cap}}$ and $\sigma_{\mathrm{cap}}$ are computed over
$\{r_j^{\mathrm{cap}}\}_{j=1}^{N}$. For localization, the $K$ trajectories
conditioned on the same parent caption form their own group:
\begin{equation}
    A_{i,k}^{\mathrm{loc}}
    =\frac{r_{i,k}^{\mathrm{loc}}-\mu_{i,\mathrm{loc}}}
    {\sigma_{i,\mathrm{loc}}+\delta},
    \label{eq:localization_advantage}
\end{equation}
with statistics taken over $k=1,\ldots,K$. This grouping preserves the
semantics of both comparisons: descriptions compete on their ability to close
the cycle, whereas localization samples compete on localization accuracy under a
fixed description.

For either task $d\in\{\mathrm{cap},\mathrm{loc}\}$, let $y$ denote a sampled
response and $A_y^d$ its group-relative advantage. For the task-specific input
$q$, the token-level importance ratio is
\begin{equation}
    \rho_{y,t}=\frac{\pi_\theta(y_t\mid q,y_{<t})}
    {\pi_{\theta_{\mathrm{old}}}(y_t\mid q,y_{<t})}.
\end{equation}
The same response-level advantage is used for every token in $y$.
We optimize the clipped GRPO objective~\cite{grpo} over valid response tokens:
\begin{equation}
\begin{aligned}
    \mathcal{J}_{d}(\theta)
    &=\mathbb{E}\!\left[
      \frac{\sum_{y\in\mathcal{G}_d}\sum_{t=1}^{|y|}\ell_d(y,t)}
      {\sum_{y\in\mathcal{G}_d}|y|}
      -\beta\mathcal{D}_{\mathrm{KL}}\right],\\
    \ell_d(y,t)
    &=\min\!\left\{\rho_{y,t} A_y^d,
      \operatorname{clip}(\rho_{y,t},1-\epsilon_l,1+\epsilon_h)A_y^d\right\}.
\end{aligned}
    \label{eq:cycle_grpo_objective}
\end{equation}
Here, $|y|$ counts valid response tokens, excluding prompt and padding tokens.
The KL regularizer against $\pi_{\mathrm{ref}}$ is estimated at token level
and averaged over the same valid tokens.
For the temporal and spatio-temporal experiments, we additionally use dual
clipping: when $A_y^d<0$, $\ell_d(y,t)$ in
Eq.~\ref{eq:cycle_grpo_objective} is replaced by
$\max\{\ell_d(y,t),cA_y^d\}$, where $c>1$.
The clipping parameters are provided in the appendix.
The captioning and localization batches share model parameters. Their gradients are
accumulated before one optimizer step, using
\begin{equation}
    \mathcal{J}_{\mathrm{CycleGRPO}}
    =\lambda_{\mathrm{cap}}\mathcal{J}_{\mathrm{cap}}
    +\lambda_{\mathrm{loc}}\mathcal{J}_{\mathrm{loc}}.
    \label{eq:joint_objective}
\end{equation}
We use equal task weights unless stated otherwise. Thus localization is not merely
a frozen scoring module: its sampled trajectories are also optimized against
the known region. Their aggregate accuracy, in turn, supplies a reference-free reward for region understanding.

\subsection{Instantiations across Region Spaces}
Generalized CycleGRPO changes only the region interface and consistency
function across domains; Eqs.~\ref{eq:caption_rollout}--\ref{eq:joint_objective}
remain unchanged.

\noindent\textbf{Spatial regions.}
For an $H\times W$ image, $\mathcal{R}\in\{0,1\}^{H\times W}$ is a mask or
$\mathcal{R}\in[0,1]^4$ a normalized box.
Our mask-based implementation uses SAMTok~\cite{zhou2026samtok} to quantize a
mask into a compact sequence in the MLLM vocabulary, allowing the same policy
to consume and emit masks without an external segmentation head. 
Mask-token agreement provides an efficient training-time consistency surrogate; decoded
mask IoU or box IoU can be used when the representation exposes geometry
directly. 
This recovers the spatial CycleGRPO formulation of our conference
work~\cite{zhang2026cyclegrpo}.

\noindent\textbf{Temporal regions.}
For a video of duration $T$, $\mathcal{R}\in[0,T]^{L\times2}$ stacks $L$
intervals $(t_\ell^{\mathrm{s}},t_\ell^{\mathrm{e}})$.
The model consumes and emits timestamp strings, and consistency is temporal IoU:
\begin{equation}
    S_{\mathrm{T}}(\mathcal{R},\widehat{\mathcal{R}})
    =\frac{\left|\mathcal{R}\cap\widehat{\mathcal{R}}\right|}
    {\left|\mathcal{R}\cup\widehat{\mathcal{R}}\right|},
    \label{eq:temporal_iou}
\end{equation}
where $|\cdot|$ denotes temporal duration and overlapping intervals are merged
before measurement. 
The outer mapping describes the content of a designated interval; the inner mapping searches the full video for that description. 
The cycle therefore couples temporal captioning and temporal localization without
using a reference caption during reinforcement learning.

\noindent\textbf{Spatio-temporal regions.}
Our video experiments represent sparse evidence as
$\mathcal{R}=\{(t_a,S_a)\}_{a=1}^{A}$ and
$\widehat{\mathcal{R}}=\{(\widehat{t}_b,\widehat{S}_b)\}_{b=1}^{B}$, where
each $S$ is a normalized box, giving tensor shapes $A\times5$ and $B\times5$.
Let $\mathcal{M}^{*}$ be the maximum-weight one-to-one matching between predicted and target anchors, where the pair weight is $|S_a\cap\widehat{S}_b|$ and is set to zero whenever $|t_a-\widehat{t}_b|>\tau$. 
We compute the joint support overlap as
\begin{equation}
\begin{aligned}
    I_{\mathrm{ST}}&=\sum_{(a,b)\in\mathcal{M}^{*}}
    |S_a\cap\widehat{S}_b|,\\
    S_{\mathrm{ST}}(\mathcal{R},\widehat{\mathcal{R}})
    &=\frac{I_{\mathrm{ST}}}
    {\sum_a|S_a|+\sum_b|\widehat{S}_b|-I_{\mathrm{ST}}}.
\end{aligned}
    \label{eq:spatiotemporal_iou}
\end{equation}
We use $\tau=3$ seconds. The one-to-one constraint prevents duplicate
predictions from reusing the same annotated evidence, while unmatched or extra
anchors remain in the union. 
Consequently, a visually similar object at the
wrong moment contributes no intersection, and a temporally aligned anchor with
an inaccurate box is also penalized.
This instantiation closes the cycle over explicit when-and-where evidence and requires structured spatio-temporal regions, but no paired textual descriptions, during
reinforcement learning.

%% file: secs/4_benchmark.tex
\section{DOVE}
\label{sec:dove}

\begin{table*}[!t]
\centering\small
\caption{Results on two inverse spatial tasks. Left: mask-to-text accuracy on
DLC-Bench~\cite{dam}. Right: text-to-mask gIoU on
GroundingSuite~\cite{zhou2026samtok}. All values are percentages.
$^{\dagger}$ denotes training on the in-domain DAM-1.5M data.}
\label{tab:main_spatial_inverse}
\label{tab:main_dlc}
\label{tab:main_groundingsuite}
\begin{minipage}[t]{0.40\textwidth}
\centering\scriptsize
\setlength{\tabcolsep}{2.2pt}
\renewcommand{\arraystretch}{1.08}
\begin{tabular}{lcccc}
\toprule
Method & Size & Pos. & Neg. & Avg. \\
\midrule
GPT-4o~\cite{gpt4o} & -- & 43.4 & 79.6 & 61.5 \\
Gemini 2.5 Pro~\cite{gemini-2.5-pro} & -- & 36.5 & 75.2 & 55.8 \\
VP-SPHINX~\cite{lin2024VP-SPHINX} & 13B & 26.3 & 71.6 & 49.0 \\
DAM$^{\dagger}$~\cite{dam} & 3B & 52.3 & 82.2 & 67.3 \\
\midrule
SAMTok~\cite{zhou2026samtok} & 4B & 43.5 & 80.4 & 61.9 \\
\rowcolor{blue!8}\textbf{\quad + Ours} & 4B & \textbf{52.4} & \textbf{83.2} & \textbf{67.8} \\
\rowcolor{blue!3}\multicolumn{2}{l}{\scriptsize $\Delta$ vs. SAMTok} & {\scriptsize +8.9} & {\scriptsize +2.8} & {\scriptsize +5.9} \\
\bottomrule
\end{tabular}
\end{minipage}
\hfill
\begin{minipage}[t]{0.56\textwidth}
\centering\scriptsize
\setlength{\tabcolsep}{2.2pt}
\renewcommand{\arraystretch}{1.08}
\begin{tabular}{lcccccc}
\toprule
Method & Size & Stuff & Part & Multi & Single & All \\
\midrule
LISA~\cite{lisa} & 7B & 85.2 & 21.2 & 71.5 & 42.8 & 57.6 \\
GLaMM~\cite{glamm} & 7B & 86.9 & 16.5 & 70.4 & 42.1 & 57.2 \\
EVF-SAM~\cite{zhang2024evf-sam} & 1B & 85.1 & 23.1 & 72.1 & 54.5 & 62.6 \\
InstructSeg~\cite{wei2025instructseg} & 3B & 56.2 & 24.2 & 66.8 & 51.3 & 52.5 \\
\midrule
SAMTok~\cite{zhou2026samtok} & 4B & 80.9 & 12.4 & 62.0 & 52.9 & 57.5 \\
\rowcolor{blue!8}\textbf{\quad + Ours} & 4B & \textbf{90.5} & \textbf{21.2} & \textbf{78.3} & \textbf{62.3} & \textbf{68.2} \\
\rowcolor{blue!3}\multicolumn{2}{l}{\scriptsize $\Delta$ vs. SAMTok} & {\scriptsize +9.6} & {\scriptsize +8.8} & {\scriptsize +16.3} & {\scriptsize +9.4} & {\scriptsize +10.7} \\
\bottomrule
\end{tabular}
\end{minipage}
\end{table*}

\begin{table}[!t]
\centering\small
\caption{Results on GAR-Bench-VQA~\cite{wang2025grasp}. We report accuracy
(\%) for fine-grained regional perception and reasoning. $^{\dagger}$ denotes
thinking mode.}
\label{tab:main_gar}
\setlength{\tabcolsep}{3pt}
\resizebox{0.8\columnwidth}{!}{%
\begin{tabular}{lccccccccc}
\toprule
\multirow{2}{*}{Method} & \multirow{2}{*}{Size} & \multirow{2}{*}{Overall}
& \multicolumn{4}{c}{Perception} & \multicolumn{3}{c}{Reasoning} \\
\cmidrule(lr){4-7}\cmidrule(lr){8-10}
& & & Color & Shape & Texture & Material & Position & Non-Entity & Relation \\
\midrule
GPT-4o~\cite{gpt4o} & -- & 53.5 & 34.8 & 65.3 & 48.3 & 52.8 & 57.8 & 60.2 & 61.4 \\
o3$^{\dagger}$~\cite{o3} & -- & 61.3 & 58.0 & 70.3 & 55.2 & 63.9 & 54.7 & 49.2 & 71.3 \\
Gemini-2.5-Pro$^{\dagger}$~\cite{gemini-2.5-pro} & -- & 64.2 & 62.3 & 68.8 & 58.6 & 66.7 & 64.1 & 64.9 & 70.3 \\
\midrule
Qwen2.5-VL~\cite{qwen25vl} & 3B & 34.4 & 29.0 & 25.0 & 34.5 & 30.6 & 43.8 & 26.2 & 44.6 \\
Qwen2.5-VL~\cite{qwen25vl} & 7B & 41.7 & 39.1 & 40.6 & 44.8 & 27.8 & 59.4 & 36.1 & 40.6 \\
Qwen2.5-VL~\cite{qwen25vl} & 32B & 50.9 & 46.4 & 53.1 & 41.4 & 30.6 & 71.9 & 36.1 & 58.4 \\
InternVL3~\cite{internvl3} & 2B & 35.1 & 30.4 & 21.9 & 48.3 & 38.9 & 48.4 & 26.2 & 38.6 \\
InternVL3~\cite{internvl3} & 8B & 38.9 & 36.2 & 37.5 & 58.6 & 41.7 & 51.6 & 27.9 & 33.6 \\
InternVL3~\cite{internvl3} & 32B & 46.5 & 39.1 & 40.6 & 51.7 & 55.6 & 60.9 & 36.1 & 47.5 \\\midrule
Sa2VA~\cite{sa2va} & 8B & 34.3 & 39.1 & 45.3 & 29.6 & 30.6 & 54.7 & 21.3 & 21.8 \\
VP-SPHINX~\cite{lin2024VP-SPHINX} & 13B & 37.5 & 33.3 & 25.0 & 44.8 & 38.9 & 60.9 & 34.3 & 32.7 \\
DAM~\cite{dam} & 3B & 38.2 & 55.1 & 39.1 & 41.4 & 36.1 & 31.3 & 36.1 & 31.7 \\
GAR~\cite{wang2025grasp} & 1B & 50.6 & 55.1 & 46.9 & 69.0 & 47.2 & 21.9 & 62.3 & 56.4 \\
GAR~\cite{wang2025grasp} & 8B & 59.9 & 59.4 & 54.7 & 75.9 & 52.8 & 48.4 & 60.7 & 68.3 \\
\midrule
SAMTok~\cite{zhou2026samtok} & 4B & 64.2 & 58.0 & 48.4 & 48.3 & 58.3 & 76.6 & 54.1 & 83.2 \\
\rowcolor{blue!8}\textbf{\quad + Ours} & 4B & \textbf{64.9} & \textbf{60.9} & \textbf{50.0} & 48.3 & \textbf{61.1} & 73.4 & 54.1 & \textbf{84.2} \\
\rowcolor{blue!3}\multicolumn{2}{l}{\scriptsize $\Delta$ vs. SAMTok} & {\scriptsize +0.7} & {\scriptsize +2.9} & {\scriptsize +1.6} & {\scriptsize +0.0} & {\scriptsize +2.8} & {\scriptsize -3.2} & {\scriptsize +0.0} & {\scriptsize +1.0} \\
\bottomrule
\end{tabular}}
\end{table}

\subsection{Task Definition}
Existing temporal-captioning benchmarks typically compare a generated caption
with one or more references. This can reward a generally plausible description
that fails to identify the designated moment, especially when the same person,
action, or scene recurs in a long video. DOVE (Distinctive Occurrence-level
Video-caption Evaluation) instead
asks whether a caption preserves the temporal evidence that identifies one
specified occurrence among visually similar events. Each benchmark group is a
tuple $\mathcal{G}=(V,q,\{S_i\}_{i=1}^{K})$, where $V$ is a source video, $q$
is a shared natural-language query, and the $K$ non-overlapping intervals are
distinct occurrences of $q$. For every target $S_i$, the remaining intervals
are same-video, same-query sibling distractors. DOVE is method-agnostic and
complements one-to-many temporal grounding~\cite{xu2026omtg}: grounding finds
the occurrences, whereas DOVE tests whether a description preserves what makes
one occurrence different from the others.

\subsection{Benchmark Construction}

\noindent
\textbf{Source collection and structural filtering.}
We draw candidate groups in approximately equal proportions from Cosmos-Cap,
Moment-10M~\cite{qian2024momentor}, and VTimeLLM~\cite{huang2024vtimellm}. We
remove source videos appearing in our training or existing test sets before
annotation. A candidate must contain two to five valid intervals for the same
query. We discard malformed and duplicate intervals, intervals covering more
than half of the video, and near-duplicate pairs whose temporal IoU exceeds
$0.5$ or for which either interval is more than $80\%$ contained in the other.
We retain one query per video, selecting the query with the largest number of valid
occurrences and breaking ties by total annotated duration. This produces a
balanced pool of 800 candidate groups (2,267 intervals), with 267/267/266
groups from the three sources.

\noindent
\textbf{Gemini pre-labeling and human review.}
We use Gemini 3 Pro~\cite{gemini} to propose, but not certify, per-interval
annotations. It receives timestamped frames sampled inside every candidate
interval together with sparse context frames from the full video, the shared
query, and all interval boundaries. For each interval, it predicts whether the
queried event occurs, proposes positive facts tagged as \emph{shared} or
\emph{specific}, and generates plausible-but-absent negatives. Providing all
intervals jointly is essential because specificity is defined within a group.
Human verification involves multiple
rounds of review and correction: reviewers first verify the occurrences,
interval distinctness, facts, tags, and negatives, and then re-examine each
retained group jointly to resolve cross-interval inconsistencies, revise
shared/specific assignments, and remove ambiguous or unsupported items. Cases
flagged during subsequent probe construction are returned for further
group-level review and correction. A group is retained only when at least two fully
reviewed intervals remain after these iterations.

\noindent
\textbf{Temporal probe construction.}
For each retained interval, we construct private multiple-choice probes along
three axes. \emph{Order} asks what happens before or after an anchor action;
\emph{State} asks how an actor or object changes; and \emph{Outcome} asks for
the visible result that identifies the occurrence. Each probe must require
evidence from multiple frames, have one unambiguous answer, and avoid
audio-dependent or purely static cues. The probes and answers undergo an
additional round of human verification and correction, including checks
against both the target interval and its sibling distractors, and are frozen
before we inspect captions from the evaluated models.

\noindent
\textbf{Benchmark statistics.}
The human-reviewed benchmark contains 180 source-video/query groups and 530
target intervals; every group comes from a distinct source video. It comprises
82 groups from Cosmos-Cap, 36 from Moment-10M, and 62 from VTimeLLM. It includes 91/41/15/33 groups with two/three/four/five repeated
occurrences, respectively. After cleaning, the annotations contain 2,602 positive
facts (1,178 shared and 1,424 specific) and 2,056 plausible negatives. The
final evaluation set contains 1,223 human-verified temporal probes: 469 Order,
420 State, and 334 Outcome questions. These probes cover 529 of the 530 target
intervals; the remaining interval has no retained probe and is excluded from
the reported metrics. Additional duration, semantic-domain, and structural
difficulty statistics are provided in Appendix~\ref{app:dove-statistics}.

\subsection{Evaluation Protocol}
Each model receives frames sampled across the full video and a prompt
specifying the target interval $S_i$. Sibling annotations,
private probes, and answers are never exposed. The model produces a free-form
caption $C_i$. A fixed text-only judge, Qwen2.5-72B-Instruct, then reads $C_i$ with each private probe
and determines whether the caption supports the correct choice, supports a
wrong choice, or does not contain enough information; the judge never sees the
video. We report temporal coverage (Temp-Cov), the fraction of probes whose correct answers are supported
by the caption; clip completeness (Clip-Complete), the fraction of intervals
for which the correct answers to all probes are supported; and separate Order, State, and Outcome
coverage, together with the not-mentioned rate (N/M). Our primary aggregate
metric is \emph{Temporal Evidence Coverage} (TEC), defined as the macro-average
of Order, State, and Outcome coverage. All methods use the same public prompt,
frozen probes, judge, and decoding protocol.

\noindent
\textbf{Statistical comparison.}
Because clips and probes within a group share one source video and same-query
distractors, we compute paired 95\% percentile bootstrap intervals by resampling
whole source-video/query groups (20,000 replicates), never individual clips or
probes. For the primary comparison against the shared Qwen3-VL initialization,
CycleGRPO improves TEC by 18.5 percentage points (95\% CI: 15.8--21.3), Temp-Cov by 18.7
points (15.9--21.5), and Clip-Complete by 9.1 points (6.4--11.9).

%% file: secs/4_exp.tex
\begin{table*}[!t]
\centering\small
\caption{Results on interleaved text-mask generation on
GCG~\cite{glamm} (left) and generalized referring expression segmentation on
GRES~\cite{gres} (right). GCG entries are Val / Test. For GRES, split entries
are Val / Test A / Test B and averages are reported separately; N-acc denotes
no-target accuracy.}
\label{tab:main_gcg}
\label{tab:main_gres}
\begin{minipage}[t]{0.46\textwidth}
\centering\scriptsize
\setlength{\tabcolsep}{1.5pt}
\renewcommand{\arraystretch}{1.0}
\resizebox{\linewidth}{!}{%
\begin{tabular}{lcccccc}
\toprule
\multirow{2}{*}{Method} & \multirow{2}{*}{Size} & METEOR & CIDEr & AP$_{50}$ & mIoU & Recall \\
\cmidrule(lr){3-7}
& & Val / Test & Val / Test & Val / Test & Val / Test & Val / Test \\
\midrule
LISA~\cite{lisa} & 7B & 13.0 / 12.9 & 33.9 / 32.2 & 25.2 / 24.8 & 62.0 / 61.7 & 36.3 / 35.5 \\
GLaMM~\cite{glamm} & 7B & 16.2 / 15.8 & 47.2 / 43.5 & 30.8 / 29.2 & 66.3 / 65.6 & 41.8 / 40.8 \\
OMG-LLaVA~\cite{zhang2024omg} & 7B & 14.9 / 14.5 & 41.2 / 38.5 & 29.9 / 28.6 & 65.6 / 64.7 & -- / -- \\
Sa2VA~\cite{sa2va} & 8B & 16.4 / 16.2 & 49.5 / 49.0 & 33.2 / 32.2 & 67.7 / 66.8 & 45.1 / 44.5 \\
\midrule
SAMTok~\cite{zhou2026samtok} & 4B & 16.1 / 16.4 & 48.2 / 51.4 & 34.7 / 34.4 & 69.4 / 68.4 & 46.6 / 48.3 \\
\rowcolor{blue!8}\textbf{\quad + Ours} & 4B & \textbf{17.3 / 17.2} & \textbf{54.3 / 53.7} & \textbf{36.8 / 35.0} & \textbf{70.2 / 69.2} & \textbf{50.2 / 49.8} \\
\rowcolor{blue!3}\multicolumn{2}{l}{\scriptsize $\Delta$ vs. SAMTok} & {\scriptsize +1.2 / +0.8} & {\scriptsize +6.1 / +2.3} & {\scriptsize +2.1 / +0.6} & {\scriptsize +0.8 / +0.8} & {\scriptsize +3.6 / +1.5} \\
\bottomrule
\end{tabular}}
\end{minipage}
\hfill
\begin{minipage}[t]{0.52\textwidth}
\centering\scriptsize
\setlength{\tabcolsep}{1.2pt}
\renewcommand{\arraystretch}{1.0}
\resizebox{\linewidth}{!}{%
\begin{tabular}{lccccccc}
\toprule
\multirow{2}{*}{Method} & \multirow{2}{*}{Size} & \multicolumn{3}{c}{Val / Test A / Test B} & \multicolumn{3}{c}{Avg.} \\
\cmidrule(lr){3-5}\cmidrule(lr){6-8}
& & gIoU & cIoU & N-acc & gIoU & cIoU & N-acc \\
\midrule
LISA~\cite{lisa} & 7B & 61.6 / 66.3 / 58.8 & 61.8 / 68.5 / 60.6 & 54.7 / 50.0 / 51.9 & 62.2 & 63.6 & 52.2 \\
SAM4MLLM~\cite{chen2024sam4mllm} & 8B & 71.9 / 74.2 / 65.3 & 67.8 / 72.2 / 63.4 & 66.1 / 63.9 / 60.0 & 70.5 & 67.8 & 63.3 \\
MLLMSeg~\cite{wang2025MLLMSeg} & 8B & 75.1 / 77.0 / 69.7 & 71.6 / 76.9 / 68.5 & 73.2 / 72.4 / 65.5 & 73.9 & 72.3 & 70.4 \\
HiMTok~\cite{himtok} & 8B & 72.1 / 73.5 / 71.7 & 70.4 / 74.9 / 72.0 & -- / -- / -- & 72.4 & 72.4 & -- \\
ARGenSeg~\cite{argenseg} & 8B & 74.7 / 73.7 / 72.4 & 72.2 / 73.6 / 70.4 & -- / -- / -- & 73.6 & 72.1 & -- \\
\midrule
SAMTok~\cite{zhou2026samtok} & 4B & 71.3 / 75.3 / 66.9 & 69.2 / 75.4 / 66.0 & 61.4 / 59.0 / 55.6 & 71.2 & 70.2 & 58.7 \\
\rowcolor{blue!8}\textbf{\quad + Ours} & 4B & \textbf{82.2 / 80.3 / 73.5} & \textbf{74.8 / 78.2 / 70.2} & \textbf{94.7 / 93.0 / 89.9} & \textbf{78.7} & \textbf{74.4} & \textbf{92.5} \\
\rowcolor{blue!3}\multicolumn{2}{l}{\scriptsize $\Delta$ vs. SAMTok} & {\scriptsize +10.9 / +5.0 / +6.6} & {\scriptsize +5.6 / +2.8 / +4.2} & {\scriptsize +33.3 / +34.0 / +34.3} & {\scriptsize +7.5} & {\scriptsize +4.2} & {\scriptsize +33.8} \\
\bottomrule
\end{tabular}}
\end{minipage}
\end{table*}

\begin{table*}[!t]
\centering\small
\caption{One-to-one temporal grounding results on
Charades-STA~\cite{gao2017charades}, ActivityNet
Captions~\cite{krishna2017activitynet}, and
QVHighlights~\cite{lei2021qvhighlights}. We report Recall@1 at temporal IoU thresholds of
$0.3$, $0.5$, and $0.7$, together with temporal IoU (tIoU), averaged over examples.}
\label{tab:main_timelens}
\setlength{\tabcolsep}{1.5pt}
\resizebox{0.8\textwidth}{!}{%
\begin{tabular}{lc cccc cccc cccc}
\toprule
\multirow{2}{*}{Model} & \multirow{2}{*}{Size} & \multicolumn{4}{c}{Charades} & \multicolumn{4}{c}{ActivityNet} & \multicolumn{4}{c}{QVHighlights} \\
\cmidrule(lr){3-6}\cmidrule(lr){7-10}\cmidrule(lr){11-14}
& & R1@0.3 & R1@0.5 & R1@0.7 & tIoU & R1@0.3 & R1@0.5 & R1@0.7 & tIoU & R1@0.3 & R1@0.5 & R1@0.7 & tIoU \\
\midrule
GPT-4o~\cite{gpt4o} & -- & 60.6 & 44.5 & 23.5 & 41.8 & 55.2 & 41.4 & 25.8 & 40.4 & 69.0 & 54.8 & 38.5 & 52.1 \\
GPT-5~\cite{gpt5} & -- & 59.3 & 42.0 & 22.0 & 40.5 & 57.4 & 44.9 & 30.4 & 42.9 & 72.4 & 60.4 & 46.4 & 56.8 \\
Gemini-2.5-Pro~\cite{gemini2_5} & -- & 74.1 & 61.1 & 34.0 & 52.8 & 72.3 & 64.2 & 47.1 & 58.1 & 84.1 & 75.9 & 61.1 & 70.4 \\
\midrule
VideoChat-Flash~\cite{videochat_flash} & 7B & 60.2 & 37.9 & 17.8 & 39.7 & 35.5 & 21.8 & 10.5 & 24.8 & 45.2 & 30.6 & 16.7 & 32.7 \\
VideoChat-R1~\cite{videochat_r1} & 7B & 51.9 & 30.8 & 11.7 & 33.7 & 35.0 & 23.9 & 11.3 & 25.0 & 29.3 & 19.1 & 9.4 & 21.5 \\
Time-R1~\cite{wang2025timer1} & 7B & 57.9 & 32.0 & 16.9 & 36.6 & 44.8 & 31.0 & 19.0 & 33.1 & 65.8 & 51.5 & 36.1 & 49.2 \\
Qwen2.5-VL~\cite{qwen25vl} & 7B & 59.7 & 37.8 & 16.6 & 39.3 & 44.1 & 31.0 & 16.1 & 31.4 & 41.5 & 27.8 & 15.2 & 31.6 \\
TimeLens~\cite{zhang2026timelens} & 7B & 70.5 & 55.6 & 28.4 & 48.8 & 62.8 & 51.0 & 32.6 & 46.2 & 74.1 & 62.7 & 43.1 & 56.0 \\
Qwen3-VL~\cite{Qwen3-VL} & 8B & 69.2 & 53.4 & 27.5 & 48.3 & 62.1 & 51.2 & 34.4 & 46.8 & 74.2 & 64.6 & 49.3 & 59.4 \\
\midrule
Qwen3-VL~\cite{Qwen3-VL} & 4B & 67.1 & 47.5 & 24.1 & 45.6 & 62.9 & 51.0 & 34.3 & 47.2 & 77.6 & 66.0 & 50.9 & 61.7 \\
\rowcolor{blue!8}\textbf{\quad + Ours} & 4B & \textbf{74.8} & \textbf{58.6} & \textbf{30.7} & \textbf{51.9} & \textbf{63.5} & \textbf{54.1} & \textbf{36.8} & \textbf{49.2} & 75.4 & 65.9 & \textbf{51.0} & \textbf{61.8} \\
\rowcolor{blue!3}\multicolumn{2}{l}{\scriptsize $\Delta$ vs. Qwen3-VL} & {\scriptsize +7.7} & {\scriptsize +11.1} & {\scriptsize +6.6} & {\scriptsize +6.3} & {\scriptsize +0.6} & {\scriptsize +3.1} & {\scriptsize +2.5} & {\scriptsize +2.0} & {\scriptsize -2.2} & {\scriptsize -0.1} & {\scriptsize +0.1} & {\scriptsize +0.1} \\
\midrule
OMTG~\cite{xu2026omtg} & 4B & 72.0 & 58.3 & 32.0 & 50.5 & 64.7 & 54.6 & 37.0 & 46.8 & 80.9 & 70.4 & 53.5 & 62.2 \\
\rowcolor{blue!8}\textbf{\quad + Ours} & 4B & \textbf{72.9} & \textbf{60.2} & \textbf{32.7} & \textbf{51.8} & \textbf{65.4} & \textbf{55.5} & \textbf{37.6} & \textbf{49.0} & 80.4 & \textbf{70.9} & \textbf{53.7} & \textbf{64.5} \\
\rowcolor{blue!3}\multicolumn{2}{l}{\scriptsize $\Delta$ vs. OMTG} & {\scriptsize +0.9} & {\scriptsize +1.9} & {\scriptsize +0.7} & {\scriptsize +1.3} & {\scriptsize +0.7} & {\scriptsize +0.9} & {\scriptsize +0.6} & {\scriptsize +2.2} & {\scriptsize -0.5} & {\scriptsize +0.5} & {\scriptsize +0.2} & {\scriptsize +2.3} \\
\bottomrule
\end{tabular}}
\end{table*}

\begin{table*}[!t]
\centering\small
\caption{Temporal captioning results on DOVE. Temp-Cov is the percentage of probes whose correct answers are supported by the caption;
Clip-Complete is the percentage of intervals with all answers supported; and
Order, State, and Outcome report type-specific coverage. N/M denotes
unmentioned probes, while Temporal Evidence Coverage (TEC) is the macro-average
of Order, State, and Outcome coverage. All values are percentages.}
\label{tab:dove_full}
\setlength{\tabcolsep}{3.5pt}
\resizebox{0.8\textwidth}{!}{%
\begin{tabular}{lcccccccc}
\toprule
Model & Size & Temp-Cov$\uparrow$ & Clip-Complete$\uparrow$ & Order$\uparrow$ & State$\uparrow$ & Outcome$\uparrow$ & N/M$\downarrow$ & TEC$\uparrow$ \\
\midrule
GPT-5.6-sol & -- & 17.3 & 5.9 & 16.4 & 17.9 & 17.7 & 71.1 & 17.3 \\
\midrule
TRACE-Uni & -- & 2.0 & 0.6 & 1.3 & 1.9 & 3.0 & 95.7 & 2.1 \\
Sa2VA~\cite{sa2va} & 8B & 0.0 & 0.0 & 0.0 & 0.0 & 0.0 & 99.9 & 0.0 \\
VideoChat2-HD~\cite{videochat2} & 7B & 1.8 & 0.4 & 1.7 & 1.7 & 2.1 & 93.5 & 1.8 \\
OMTG~\cite{xu2026omtg} & 4B & 0.7 & 0.2 & 0.6 & 1.0 & 0.6 & 98.8 & 0.7 \\
VideoChat-R1~\cite{videochat_r1} & 7B & 3.2 & 0.4 & 2.8 & 2.9 & 4.2 & 93.9 & 3.3 \\
Time-R1~\cite{wang2025timer1} & 7B & 2.6 & 0.6 & 2.3 & 2.9 & 2.7 & 94.8 & 2.6 \\
TimeLens~\cite{zhang2026timelens} & 7B & 4.4 & 0.9 & 3.8 & 4.3 & 5.4 & 90.8 & 4.5 \\
LLaVA-Video~\cite{llava-video} & 7B & 8.3 & 2.5 & 8.3 & 8.3 & 8.1 & 81.1 & 8.2 \\
Video-LLaMA3~\cite{damonlpsg2025videollama3} & 7B & 1.0 & 0.6 & 0.4 & 1.2 & 1.5 & 97.5 & 1.0 \\
InternVL3~\cite{internvl3} & 8B & 2.4 & 0.6 & 3.0 & 1.4 & 2.7 & 94.1 & 2.4 \\
Open-o3-Video~\cite{meng2025openo3video} & 4B & 13.5 & 4.7 & 11.5 & 14.8 & 14.7 & 76.2 & 13.6 \\
\midrule
Qwen3-VL~\cite{Qwen3-VL} & 4B & 14.7 & 5.5 & 12.6 & 17.1 & 14.7 & 75.1 & 14.8 \\
\rowcolor{blue!8}\textbf{\quad + Ours} & 4B & \textbf{33.4} & \textbf{14.6} & \textbf{31.8} & \textbf{37.4} & \textbf{30.8} & \textbf{50.7} & \textbf{33.3} \\
\rowcolor{blue!3}\multicolumn{2}{l}{\scriptsize $\Delta$ vs. Qwen3-VL} & {\scriptsize +18.7} & {\scriptsize +9.1} & {\scriptsize +19.2} & {\scriptsize +20.3} & {\scriptsize +16.1} & {\scriptsize -24.4} & {\scriptsize +18.5} \\
\bottomrule
\end{tabular}}
\end{table*}

\section{Experiments}
\label{sec:exp}

\noindent
\textbf{Experimental setup.}
We evaluate the same region--text--region principle in three spaces: spatial masks, temporal intervals, and sparse spatio-temporal trajectories. 
Spatial experiments retain the published ECCV setup~\cite{zhang2026cyclegrpo} and start from Qwen3-VL-4B-SAMTok~\cite{zhou2026samtok}. 
Temporal experiments compare captioning-only GRPO, localization-only GRPO, and Generalized CycleGRPO from one shared Qwen3-VL-4B co-SFT initialization, using identical RL videos and budgets.
Temporal training data are drawn from OMTG~\cite{xu2026omtg}.
For the main spatio-temporal experiments, we initialize from Open-o3-Video-4B
(RL) after one epoch of inverse-caption and video-QA co-SFT.
The training data for these experiments are drawn from Open-o3-Video~\cite{meng2025openo3video}.
Following the multi-task training strategy of Open-o3-Video~\cite{meng2025openo3video},
we mix spatial, temporal, and spatio-temporal region tasks with general video
QA. The region tasks use Generalized CycleGRPO; the auxiliary QA task retains
answer-based rewards. The data composition and training settings are given in
Tab.~\ref{tab:app-st-settings}.
We apply early stopping based on training-reward convergence.
We report complete configurations in the appendix.

\noindent
\textbf{Datasets and benchmarks.}
Spatial evaluation uses DLC-Bench~\cite{dam},
GroundingSuite~\cite{zhou2026samtok}, GAR-Bench-VQA~\cite{wang2025grasp}, GCG~\cite{glamm}, and GRES~\cite{gres}, following their official metrics.
Temporal localization is evaluated on TimeLens~\cite{zhang2026timelens} and TVGBench~\cite{wang2025timer1}; joint temporal--spatial reasoning is evaluated
on V-STaR~\cite{cheng2025vstar}; and interval-conditioned captioning is evaluated on DOVE (Sec.~\ref{sec:dove}).
Besides general-purpose MLLMs and the ECCV spatial baselines, we include OMTG~\cite{xu2026omtg} and Open-o3-Video~\cite{meng2025openo3video} as specialized temporal and spatio-temporal comparisons, respectively.

\noindent
\textbf{Implementation details.}
Our training runs use bfloat16 and gradient checkpointing with a frozen visual encoder.
We train GRPO variants with our extended \texttt{verl} codebase and FSDP. 
The appendix provides per-setting data, optimization, rollout, and input configurations.

\noindent
\textbf{Table conventions.} Blue rows identify our variants, and rows labeled $\Delta$ report signed changes relative to the indicated base.

\subsection{Main Results}

\noindent
\textbf{Mask-to-text on DLC-Bench.}
Tab.~\ref{tab:main_dlc} evaluates whether a model can describe the content of
a designated mask. CycleGRPO obtains 67.8\% average accuracy, up from 61.9\%
for the SAMTok initialization. It also slightly exceeds DAM (67.3\%),
although DAM uses the in-domain DAM-1.5M training set. The main gain occurs on
positive probes: CycleGRPO reaches 52.4\%, compared with 43.5\% for SAMTok,
while also improving negative-probe accuracy from 80.4\% to 83.2\%. The larger
gain on positive probes suggests that closing the cycle strengthens recognition of content inside the queried mask.

\noindent
\textbf{Text-to-mask on GroundingSuite.}
Tab.~\ref{tab:main_groundingsuite} tests the inverse mapping from a referring
expression to a mask. CycleGRPO raises overall gIoU from 57.5\% to 68.2\%. The gains cover every subset and are largest on multi-object expressions, indicating that the cycle is particularly helpful when grounding requires resolving object identity or grouping, where
an underspecified description is most likely to reconstruct the wrong mask.

\noindent
\textbf{Fine-grained regional understanding on GAR-Bench-VQA.}
As shown in Tab.~\ref{tab:main_gar}, CycleGRPO achieves the highest overall
accuracy of 64.9\%, compared with 64.2\% for both SAMTok and Gemini-2.5-Pro. 
The overall gain is modest and not uniform across categories. Relative to SAMTok,
CycleGRPO improves color, shape, and material recognition from
58.0/48.4/58.3\% to 60.9/50.0/61.1\%, and achieves the highest listed relation accuracy of
84.2\%, but position accuracy decreases from 76.6\% to 73.4\% and non-entity
accuracy remains 54.1\%. These results suggest that the cycle improves target-specific appearance and relation understanding, but does not improve every aspect of spatial reasoning.

\noindent
\textbf{Interleaved generation and localization on GCG.}
Tab.~\ref{tab:main_gcg} evaluates language generation and mask prediction in
the same output. CycleGRPO improves all ten reported validation/test metrics
over SAMTok. Validation CIDEr/mIoU/Recall increase from 48.2/69.4/46.6 to
54.3/70.2/50.2, and test CIDEr/mIoU/Recall increase from 51.4/68.4/48.3 to
53.7/69.2/49.8. AP$_{50}$ also rises from 34.7/34.4 to 36.8/35.0 on the two
splits. The simultaneous gains show that language quality and mask localization improve together.

\noindent
\textbf{Generalized referring segmentation on GRES.}
Tab.~\ref{tab:main_gres} further tests localization with single, multiple, and
absent targets. CycleGRPO improves average gIoU/cIoU from 71.2/70.2\% for
SAMTok to 78.7/74.4\%. Its gIoU is consistently higher on validation, Test A, and Test
B, showing that the average gain is not driven by a single split. No-target
accuracy also rises from 58.7\% to 92.5\% on average. The
complete cycle therefore improves both rejection of absent targets and
localization of targets that are present.

\noindent
\textbf{Generalization to continuous bounding boxes.}
The spatial instantiation is not tied to discrete mask tokens. We replace the
mask representation with continuous bounding-box coordinates and apply the
same region--text--region objective to three MLLM backbones.
Tab.~\ref{tab:app-spatial-bbox} shows consistent improvements on DLC-Bench and
GroundingSuite across model families and scales. 
In particular, the overall
GroundingSuite accuracy increases from 69.0\% to 69.9\% for Qwen2.5-VL-3B and
from 71.4\% to 74.1\% for both Qwen3-VL-4B and Qwen3-VL-8B.

\begin{table}[!t]
\centering\small
\captionof{table}{Ablation on spatial localization formats using continuous
bounding-box coordinates. Results are reported on DLC-Bench~\cite{dam} and
GroundingSuite~\cite{zhou2026samtok}; all values are percentages.}
\label{tab:app-spatial-bbox}
\setlength{\tabcolsep}{3pt}
\resizebox{0.8\columnwidth}{!}{%
\begin{tabular}{lccc|ccccc}
\toprule
\multirow{2}{*}{Method} & \multicolumn{3}{c|}{DLC-Bench} & \multicolumn{5}{c}{GroundingSuite (Acc@0.5)} \\
\cmidrule(lr){2-4}\cmidrule(lr){5-9}
& Pos. & Neg. & Avg. & Single & Stuff & Multi & Part & All \\
\midrule
Qwen2.5-VL-3B (bbox) & 19.4 & 44.8 & 32.1 & 93.7 & 32.8 & 63.2 & 66.3 & 69.0 \\
\rowcolor{blue!8}\textbf{\quad + Ours} & \textbf{23.3} & \textbf{46.6} & \textbf{35.0} & \textbf{94.1} & \textbf{34.1} & \textbf{64.1} & \textbf{67.3} & \textbf{69.9} \\
Qwen3-VL-4B (bbox) & 39.6 & 62.4 & 51.0 & \textbf{95.2} & 24.2 & \textbf{76.7} & 67.0 & 71.4 \\
\rowcolor{blue!8}\textbf{\quad + Ours} & \textbf{43.2} & \textbf{65.4} & \textbf{54.3} & 94.9 & \textbf{33.9} & 76.3 & \textbf{71.1} & \textbf{74.1} \\
Qwen3-VL-8B (bbox) & 41.0 & \textbf{65.4} & 53.2 & 94.9 & 31.2 & 71.8 & 67.4 & 71.4 \\
\rowcolor{blue!8}\textbf{\quad + Ours} & \textbf{49.4} & 65.0 & \textbf{57.2} & \textbf{95.9} & \textbf{37.6} & \textbf{74.4} & \textbf{70.4} & \textbf{74.1} \\
\bottomrule
\end{tabular}}
\end{table}

\noindent
\textbf{Temporal grounding on TimeLens.}
Tab.~\ref{tab:main_timelens} reveals that the effect of CycleGRPO depends on
the temporal initialization. Starting from Qwen3-VL-4B-Instruct, CycleGRPO
improves all four Charades metrics: R1@$0.3/0.5/0.7$ rises from
67.1/47.5/24.1\% to 74.8/58.6/30.7\%, and tIoU from 45.6\% to 51.9\%. All four
ActivityNet metrics also improve, including tIoU from 47.2\% to 49.2\%. On
QVHighlights, performance is largely preserved: R1@$0.3/0.5$ changes from
77.6/66.0\% to 75.4/65.9\%, while R1@$0.7$ and tIoU increase slightly from
50.9/61.7\% to 51.0/61.8\%. Overall, the Qwen-initialized policy improves 10 of 12
entries, with its largest gains on Charades.

With the specialized OMTG-4B (RL) initialization, the gains are
more consistent across datasets. CycleGRPO improves 11 of the 12 entries: tIoU increases from
50.5\% to 51.8\% on Charades, 46.8\% to 49.0\% on ActivityNet, and 62.2\% to
64.5\% on QVHighlights. On QVHighlights, R1@$0.5$ and R1@$0.7$ rise from
70.4/53.5\% to 70.9/53.7\%, while R1@$0.3$ is the only regression (80.9\% to 80.4\%).
Together, the two initialization settings show that the cyclic objective
improves temporal grounding broadly, while the magnitude and dataset balance
of the gains still depend on the initial policy.

\noindent
\textbf{Region-conditioned captioning on DOVE.}
Tab.~\ref{tab:dove_full} evaluates whether a description
is specific enough to identify its designated occurrence among similar clips.
The CycleGRPO model, also evaluated in Tab.~\ref{tab:main_temporal_objectives}, improves over its Qwen3-VL-4B-Instruct initialization from 14.8\% to 33.3\% TEC, from 14.7\% to 33.4\% in temporal coverage, and from 5.5\% to 14.6\% in clip completeness.
It also outperforms Open-o3-Video-4B (RL), which reaches 13.6\% TEC, with large gains in Order (31.8\% versus 11.5\%) and Outcome (30.8\% versus 14.7\%), and exceeds GPT-5.6-sol's 17.3\% TEC.
Relative to the Qwen initialization, N/M decreases from 75.1\% to 50.7\%, showing that the generated captions convey substantially more target-specific evidence.
These results show improved coverage of target-specific temporal evidence across all three probe types.

\begin{table}[!t]
\centering\small
\caption{Results on V-STaR~\cite{cheng2025vstar}. \textit{What} measures answer
accuracy, \textit{When} measures temporal IoU, and \textit{Where} measures visual IoU. 
Chain1 denotes what--when--where and Chain2 denotes what--where--when; mAM and mLGM
are the aggregate metrics.}
\label{tab:main_vstar}
\setlength{\tabcolsep}{5pt}
\resizebox{0.8\columnwidth}{!}{%
\begin{tabular}{lcccccccc}
\toprule
\multirow{2}{*}{Model} & \multirow{2}{*}{Size} & What & \multicolumn{2}{c}{When (Temporal IoU)} & \multicolumn{2}{c}{Where (Visual IoU)} & \multicolumn{2}{c}{Overall} \\
\cmidrule(lr){3-3}\cmidrule(lr){4-5}\cmidrule(lr){6-7}\cmidrule(lr){8-9}
& & Acc & Chain1 & Chain2 & Chain1 & Chain2 & mAM & mLGM \\
\midrule
GPT-4o~\cite{gpt4o} & -- & 60.8 & 16.7 & 12.8 & 6.5 & 3.0 & 26.8 & 38.2 \\
Gemini-2-Flash~\cite{gemini} & -- & 53.0 & 24.5 & 23.8 & 4.6 & 2.2 & 26.9 & 35.6 \\
\midrule
Video-LLaMA3~\cite{damonlpsg2025videollama3} & 7B & 41.9 & 23.0 & 23.1 & 0.9 & 0.2 & 21.7 & 27.0 \\
LLaVA-Video~\cite{llava-video} & 7B & 49.5 & 10.5 & 12.2 & 1.9 & 1.3 & 20.8 & 27.3 \\
VideoChat2~\cite{videochat2} & 7B & 36.2 & 13.7 & 12.5 & 2.5 & 1.0 & 17.0 & 20.3 \\
Oryx-1.5~\cite{liu2024oryx} & 7B & 20.5 & 13.5 & 14.8 & 10.1 & 3.5 & 15.1 & 13.8 \\
InternVL-2.5~\cite{internvl2.5} & 8B & 44.2 & 8.7 & 7.8 & 0.7 & 0.1 & 17.6 & 24.9 \\
Qwen2.5-VL (base)~\cite{qwen25vl} & 7B & 33.5 & 15.4 & 13.8 & 17.0 & 2.5 & 19.3 & 22.4 \\
TRACE~\cite{guo2025trace} & 7B & 17.6 & 19.1 & 17.1 & 0.0 & 0.0 & 12.0 & 13.3 \\
Sa2VA~\cite{sa2va} & 8B & 16.4 & 0.1 & 0.0 & 32.3 & 37.5 & 17.1 & 20.3 \\
Open-o3-Video~\cite{meng2025openo3video} & 7B & 61.0 & 24.5 & 24.0 & 25.4 & 6.0 & 33.7 & 46.6 \\
\midrule
Open-o3-Video~\cite{meng2025openo3video} & 4B & 59.5 & 25.2 & 23.7 & 31.5 & 8.2 & 34.6 & 47.2 \\
\rowcolor{blue!8}\textbf{\quad + Ours} & 4B & \textbf{60.7} & \textbf{26.2} & \textbf{25.8} & \textbf{33.2} & \textbf{10.5} & \textbf{36.2} & \textbf{49.8} \\
\rowcolor{blue!3}\multicolumn{2}{l}{\scriptsize $\Delta$ vs. Open-o3-Video} & {\scriptsize +1.2} & {\scriptsize +1.0} & {\scriptsize +2.1} & {\scriptsize +1.7} & {\scriptsize +2.3} & {\scriptsize +1.6} & {\scriptsize +2.6} \\
\bottomrule
\end{tabular}}
\end{table}

\noindent
\textbf{Joint when-and-where reasoning on V-STaR.}
Tab.~\ref{tab:main_vstar} shows coordinated gains across all components of
spatio-temporal grounding. Relative to the Open-o3-Video-4B (RL) checkpoint,
Generalized CycleGRPO increases answer accuracy from 59.5\% to 60.7\%; temporal IoU from
25.2/23.7\% to 26.2/25.8\% for the two reasoning chains; and visual IoU from
31.5/8.2\% to 33.2/10.5\%. Consequently, mAM and mLGM improve from 34.6/47.2\%
to 36.2/49.8\%. Generalized CycleGRPO also exceeds the reported Open-o3-Video-7B model in
both aggregate metrics (33.7/46.6\%), despite using a 4B policy. The simultaneous
improvement in answer, temporal, and spatial measures supports optimizing the
joint trajectory rather than either of its projections alone.

\begin{table*}[!tp]
\centering\small
\caption{Results across general video understanding, reasoning, and temporal
grounding. VideoMME~\cite{fu2024videomme},
WorldSense~\cite{hong2025worldsense}, and
VideoMMMU~\cite{hu2025videommmu} report accuracy, while
TVGBench~\cite{wang2025timer1} reports mIoU. Avg. is the mean over the reported
columns.}
\label{tab:main_general_video}
\setlength{\tabcolsep}{3.5pt}
\resizebox{0.7\textwidth}{!}{%
\begin{tabular}{lccccccccc}
\toprule
\multirow{2}{*}{Model} & \multirow{2}{*}{Size} & \multicolumn{2}{c}{VideoMME} & \multicolumn{2}{c}{WorldSense} & \multicolumn{2}{c}{VideoMMMU} & TVGBench & \multirow{2}{*}{Avg.} \\
\cmidrule(lr){3-4}\cmidrule(lr){5-6}\cmidrule(lr){7-8}\cmidrule(lr){9-9}
& & Overall & Long & Overall & Recognition & Overall & Perception & mIoU & \\
\midrule
GPT-4o~\cite{gpt4o} & -- & 71.9 & -- & 42.6 & -- & 61.2 & 66.0 & -- & -- \\
\midrule
Video-LLaMA3~\cite{damonlpsg2025videollama3} & 7B & 60.6 & 48.7 & 37.3 & 38.1 & 46.5 & 59.7 & 22.2 & 44.7 \\
InternVL-2.5~\cite{internvl2.5} & 8B & 62.3 & 51.2 & 39.6 & 38.5 & 42.4 & 57.0 & 6.3 & 42.5 \\
Qwen2.5-VL (base)~\cite{qwen25vl} & 7B & 62.4 & 50.8 & 36.1 & 33.7 & 51.2 & 64.7 & 16.3 & 45.0 \\
VideoRFT~\cite{wang2025videorft} & 7B & 59.8 & 50.7 & 38.2 & 36.6 & 51.1 & 66.0 & 14.3 & 45.2 \\
VideoR1~\cite{videor1} & 7B & 61.4 & 50.6 & 35.5 & 32.8 & 52.4 & 65.3 & 9.6 & 43.9 \\
\midrule
Open-o3-Video~\cite{meng2025openo3video} & 4B & 61.0 & 50.2 & 39.3 & 38.1 & 58.0 & 76.0 & 21.1 & 49.1 \\
\rowcolor{blue!8}\textbf{\quad + Ours} & 4B & \textbf{62.3} & \textbf{52.1} & \textbf{40.2} & \textbf{39.3} & \textbf{59.7} & 75.7 & \textbf{28.5} & \textbf{51.1} \\
\rowcolor{blue!3}\multicolumn{2}{l}{\scriptsize $\Delta$ vs. Open-o3-Video} & {\scriptsize +1.3} & {\scriptsize +1.9} & {\scriptsize +0.9} & {\scriptsize +1.2} & {\scriptsize +1.7} & {\scriptsize -0.3} & {\scriptsize +7.4} & {\scriptsize +2.0} \\
\bottomrule
\end{tabular}}
\end{table*}
\noindent
\textbf{General video capabilities and temporal grounding.}
Tab.~\ref{tab:main_general_video} checks whether the V-STaR gains are obtained
at the expense of broader video understanding. Generalized CycleGRPO improves the local
Open-o3-Video-4B (RL) checkpoint on VideoMME Overall/Long from 61.0/50.2\% to
62.3/52.1\%, WorldSense Overall/Recognition from 39.3/38.1\% to 40.2/39.3\%, and
VideoMMMU Overall from 58.0\% to 59.7\%. The largest gain is on TVGBench, where
mIoU rises from 21.1\% to 28.5\%, exceeding all listed open-source
baselines. VideoMMMU Perception is the only regression (76.0\% to 75.7\%). Averaged
over the reported columns, the score increases from 49.1\% to 51.1\%. Thus, the
trajectory-level cycle improves temporal grounding while preserving the
initial policy's broader video understanding capabilities.

\begin{figure*}[!t]
    \centering
    \includegraphics[width=0.80\textwidth]{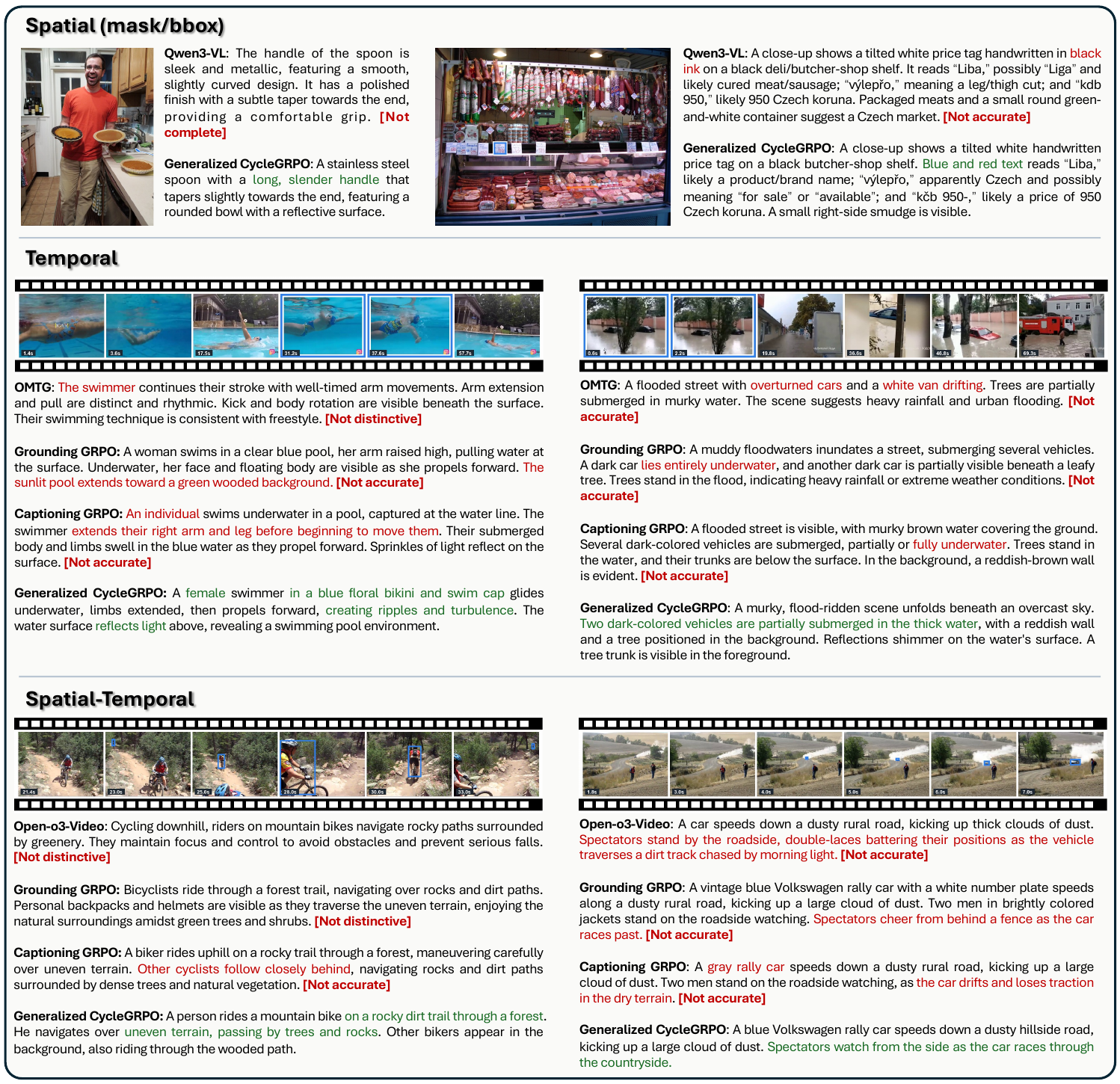}
    \caption{Region-conditioned captioning across spatial, temporal, and
    spatio-temporal regions. Blue marks targets, red marks errors or omissions,
    and green marks target-specific evidence from Generalized CycleGRPO.}
    \label{fig:visual-analysis}
\end{figure*}

\subsection{Ablation Study and Analysis}

\begin{table*}[!t]
\centering\small
\caption{Reward-side ablation for temporal grounding on
Charades-STA~\cite{gao2017charades} with OMTG~\cite{xu2026omtg}, and
temporal captioning on DOVE with Qwen3-VL~\cite{Qwen3-VL}.
All values are percentages.}
\label{tab:cycle_reward_side_ablation}
\setlength{\tabcolsep}{2.2pt}
\resizebox{0.90\textwidth}{!}{%
\begin{tabular}{lcc|cccc|ccccccc}
\toprule
\multirow{2}{*}{Model} & \multirow{2}{*}{\makecell{Caption\\Reward}} & \multirow{2}{*}{\makecell{Localization\\Reward}}
& \multicolumn{4}{c|}{Charades} & \multicolumn{7}{c}{DOVE} \\
\cmidrule(lr){4-7}\cmidrule(lr){8-14}
& & & R1@0.3 & R1@0.5 & R1@0.7 & tIoU
& TEC$\uparrow$ & Temp-Cov$\uparrow$ & Clip-Complete$\uparrow$ & Order$\uparrow$ & State$\uparrow$ & Outcome$\uparrow$ & N/M$\downarrow$ \\
\midrule
Baseline & -- & --
& 72.0 & 58.3 & 32.0 & 50.5
& 14.8 & 14.7 & 5.5 & 12.6 & 17.1 & 14.7 & 75.1 \\
\quad + Caption reward only & \cmark & \xmark
& \textbf{73.0} & 59.6 & 32.3 & 51.4
& 30.1 & 30.0 & 13.4 & 26.9 & 33.8 & 29.6 & 53.0 \\
\quad + Localization reward only & \xmark & \cmark
& 71.5 & 58.5 & 32.4 & 51.0
& 29.6 & 29.7 & 13.8 & 29.0 & 32.1 & 27.5 & 52.9 \\
\rowcolor{blue!8}\textbf{\quad + Ours} & \cmark & \cmark
& 72.9 & \textbf{60.2} & \textbf{32.7} & \textbf{51.8}
& \textbf{33.3} & \textbf{33.4} & \textbf{14.6} & \textbf{31.8} & \textbf{37.4} & \textbf{30.8} & \textbf{50.7} \\
\bottomrule
\end{tabular}}
\end{table*}

\noindent
\textbf{Why optimize both reward sides?}
Tab.~\ref{tab:cycle_reward_side_ablation} compares caption-only, localization-only,
and joint rewards while retaining the region--text--region rollout. The
Charades and DOVE comparisons use the OMTG and Qwen3-VL co-SFT
initializations, respectively. Caption-only
optimization produces the highest Charades R1@$0.3$ (73.0\%, versus 72.9\% for
the full objective), but remains lower at the stricter thresholds and in tIoU.
On DOVE, caption-only and localization-only optimization achieve TEC scores
of 30.1\% and 29.6\%, respectively. In contrast, the full Generalized CycleGRPO
objective performs best on Charades R1@$0.5/0.7$ and tIoU, and raises DOVE TEC
to 33.3\% and clip completeness to 14.6\%. Its N/M further decreases to 50.7\%,
compared with 53.0\% and 52.9\% for the two single-reward variants. At the
evaluated training points, the joint objective achieves higher temporal
coverage and clip completeness than either single-reward variant.

\begin{table*}[!tp]
\centering\small
\caption{Training-objective ablation with Qwen3-VL~\cite{Qwen3-VL} for
temporal grounding on Charades-STA~\cite{gao2017charades} and temporal
captioning on DOVE. All values are percentages.}
\label{tab:main_temporal_objectives}
\setlength{\tabcolsep}{2.2pt}
\resizebox{0.90\textwidth}{!}{%
\begin{tabular}{lcccc|ccccccc}
\toprule
\multirow{2}{*}{Model} & \multicolumn{4}{c|}{Charades} & \multicolumn{7}{c}{DOVE} \\
\cmidrule(lr){2-5}\cmidrule(lr){6-12}
& R1@0.3 & R1@0.5 & R1@0.7 & tIoU
& TEC$\uparrow$ & Temp-Cov$\uparrow$ & Clip-Complete$\uparrow$ & Order$\uparrow$ & State$\uparrow$ & Outcome$\uparrow$ & N/M$\downarrow$ \\
\midrule
Qwen3-VL~\cite{Qwen3-VL}
& 67.1 & 47.5 & 24.1 & 45.6
& 14.8 & 14.7 & 5.5 & 12.6 & 17.1 & 14.7 & 75.1 \\
\quad + Captioning-only GRPO
& 69.9 & 54.2 & 28.6 & 49.0
& 17.0 & 17.0 & 6.6 & 13.4 & 22.9 & 14.7 & 74.9 \\
\quad + Localization-only GRPO
& 74.4 & 58.0 & 29.8 & 51.2
& 29.4 & 29.3 & 13.6 & 25.8 & 33.3 & 29.0 & 52.9 \\
\rowcolor{blue!8}\textbf{\quad + Ours}
& \textbf{74.8} & \textbf{58.6} & \textbf{30.7} & \textbf{51.9}
& \textbf{33.3} & \textbf{33.4} & \textbf{14.6} & \textbf{31.8} & \textbf{37.4} & \textbf{30.8} & \textbf{50.7} \\
\bottomrule
\end{tabular}}
\end{table*}

\noindent
\textbf{Why optimize the complete cycle?}
Tab.~\ref{tab:main_temporal_objectives} controls the initialization and RL
videos while changing only the optimized mapping and reward. 
Both one-directional baselines require externally provided captions during RL: captioning-only GRPO
uses them as reference targets for a language-based reward, whereas
localization-only GRPO uses them as localization prompts. 
In contrast, CycleGRPO generates its intermediate descriptions from the annotated regions
and requires neither reference captions nor an external language judge during
RL. Captioning-only GRPO raises Charades R1@$0.3/0.5/0.7$ from 67.1/47.5/24.1\% to
69.9/54.2/28.6\% and tIoU from 45.6\% to 49.0\%. On DOVE, it modestly raises
TEC from 14.8\% to 17.0\%, but clip completeness remains low at 6.6\%.
Optimizing captioning against a reference-based judge thus improves localization as well, but yields only modest gains in distinguishing the target occurrence from same-query distractors.

Localization-only GRPO provides a stronger localization signal: it reaches 51.2\%
Charades tIoU and 29.4\% TEC on DOVE, against 45.6\% and 14.8\% for the
initialization. 
However, it never directly optimizes whether a region-conditioned description can be localized back to its source interval.
CycleGRPO explicitly optimizes this reconstruction and achieves the best joint result:
51.9\% Charades tIoU and 33.3\% TEC on DOVE.
Relative to localization-only GRPO, it raises R1@$0.3/0.5/0.7$ from 74.4/58.0/29.8\% to 74.8/58.6/30.7\%
and tIoU from 51.2\% to 51.9\%.
On DOVE, temporal coverage rises from 29.3\% to 33.4\%, clip completeness from 13.6\% to 14.6\%, and Order/State/Outcome coverage from 25.8/33.3/29.0\% to 31.8/37.4/30.8\%, while N/M decreases from
52.9\% to 50.7\%. 
The complete cycle therefore achieves the best results on both temporal localization and target-specific captioning in this comparison.

\begin{table}[!t]
\centering\small
\caption{Training-objective ablation with Open-o3-Video-4B~\cite{meng2025openo3video}
for spatio-temporal grounding on V-STaR~\cite{cheng2025vstar}, video
understanding on VideoMME~\cite{fu2024videomme}, WorldSense~\cite{hong2025worldsense},
and VideoMMMU~\cite{hu2025videommmu}, and temporal grounding on
TVGBench~\cite{wang2025timer1}. All values are percentages.}
\label{tab:openo3_objective_ablation}
\setlength{\tabcolsep}{2.6pt}
\resizebox{0.8\columnwidth}{!}{%
\begin{tabular}{lccccc}
\toprule
Model & V-STaR & VideoMME & WorldSense & VideoMMMU & TVGBench \\
& mLGM & Acc. & Acc. & Acc. & mIoU \\
\midrule
Open-o3-Video~\cite{meng2025openo3video}
& 47.2 & 61.0 & 39.3 & 58.0 & 21.1 \\
\quad + Captioning-only GRPO
& 47.9 & 61.8 & 39.1 & 58.4 & 23.7 \\
\quad + Localization-only GRPO
& 47.3 & 61.9 & 39.6 & 58.1 & 24.9 \\
\rowcolor{blue!8}\textbf{\quad + Ours}
& \textbf{49.8} & \textbf{62.3} & \textbf{40.2} & \textbf{59.7} &
\textbf{28.5} \\
\bottomrule
\end{tabular}}
\end{table}

\noindent
\textbf{Controlled objectives on Open-o3-Video.}
Tab.~\ref{tab:openo3_objective_ablation} further isolates the RL objective in
the spatio-temporal setting. 
Captioning-only and localization-only GRPO improve TVGBench mIoU from 21.1\% to 23.7\% and 24.9\%, respectively, but their gains are less consistent across the five benchmarks.
The complete cycle performs best on every reported aggregate: it reaches
49.8\% V-STaR mLGM, 62.3\% VideoMME accuracy, 40.2\% WorldSense accuracy,
59.7\% VideoMMMU accuracy, and 28.5\% TVGBench mIoU. 
These controlled results show that optimizing either direction alone does not recover the joint gains from coupling captioning and localization.

\begin{table}[!t]
\centering\small
\caption{Spatio-temporal grounding and general video understanding results
with Open-o3-Video~\cite{meng2025openo3video} based on Qwen2.5-VL-7B.
V-STaR~\cite{cheng2025vstar} reports mAM and
mLGM; VideoMME~\cite{fu2024videomme} and
WorldSense~\cite{hong2025worldsense} report overall accuracy. All values are
percentages.}
\label{tab:qwen25vl7b_cyclegrpo}
\setlength{\tabcolsep}{2pt}
\resizebox{0.5\columnwidth}{!}{%
\begin{tabular}{lcccc}
\toprule
Model & \multicolumn{2}{c}{V-STaR} & VideoMME & WorldSense \\
\cmidrule(lr){2-3}\cmidrule(lr){4-4}\cmidrule(lr){5-5}
& mAM & mLGM & Overall & Overall \\
\midrule
Open-o3-Video~\cite{meng2025openo3video}
& 33.4 & 45.7 & 63.0 & 39.0 \\
\rowcolor{blue!8}\textbf{\quad + Ours}
& \textbf{34.4} & \textbf{47.5} & \textbf{63.4} & \textbf{40.0} \\
\bottomrule
\end{tabular}}
\end{table}

\noindent
\textbf{Controlled Qwen2.5-VL-7B comparison.}
Tab.~\ref{tab:qwen25vl7b_cyclegrpo} re-evaluates the Open-o3-Video
initialization and CycleGRPO under the same protocol. 
CycleGRPO improves V-STaR mAM/mLGM from 33.4/45.7\% to 34.4/47.5\%, VideoMME Overall from 63.0\% to 63.4\%, and WorldSense Overall from 39.0\% to 40.0\%.
The consistent gains show that the cycle is not tied to the 4B policy used in
the main spatio-temporal experiments: it transfers to a Qwen2.5-VL-7B backbone
and improves both joint grounding and general video understanding under a
controlled evaluation protocol.

\noindent
\textbf{What generalizes, and what remains distribution-dependent?}
Across the three region spaces, the results support joint improvements in understanding and localization: caption quality and mask localization improve in the spatial setting (Tab.~\ref{tab:main_gcg}); Charades localization and DOVE coverage improve in the temporal setting (Tab.~\ref{tab:main_temporal_objectives}); and V-STaR shows gains in joint temporal--spatial localization (Tab.~\ref{tab:main_vstar}).
At the same time, the Qwen-initialized temporal-grounding results in Tab.~\ref{tab:main_timelens} show that cycle consistency is not a substitute for broad coverage of target video domains.
We therefore report results for each dataset and threshold to make the QVHighlights trade-offs explicit.

\subsection{Visual Analysis}

\noindent
\textbf{Captioning across region spaces.}
In Fig.~\ref{fig:visual-analysis}, the first row shows that the spatial benefit is not
limited to identifying an object category. 
For the masked spoon, Qwen3-VL describes only the handle and omits the bowl, whereas Generalized CycleGRPO captures both components and their appearance. 
For the price tag, it also recovers visible color and text details more faithfully, rather than introducing unsupported readings.

The second row of Fig.~\ref{fig:visual-analysis} presents temporal examples that require the caption to isolate the designated interval from the surrounding video.
The baselines either remain generic or introduce events outside the target, such as an extended pool background, an exaggerated limb motion, overturned vehicles, or a drifting white van. 
Generalized CycleGRPO instead retains interval-specific evidence: the swimmer's floral
bikini and cap together with the induced ripples, and two dark vehicles only
partially submerged in the floodwater.

The last row of Fig.~\ref{fig:visual-analysis} presents spatio-temporal cases in which the description must track an entity across timestamps.
Generalized CycleGRPO associates the target rider with the rocky forest trail and uneven terrain, and identifies the blue rally car, dusty hillside road, and roadside spectators without hallucinating cheering, a fence, or loss of traction. 
Together, these examples illustrate region-conditioned captioning across masks, intervals, and trajectories, with descriptions that retain target-specific visual evidence.
Additional spatial, temporal, and spatio-temporal examples, including failure
cases, are provided in Appendices~\ref{app:spatial-additional},
\ref{app:temporal-visualization}, and~\ref{app:vstar-visualization}.

\FloatBarrier

%% file: secs/5_limitations.tex
\section{Limitations and Future Work}
\label{sec:limitations}

\noindent\textbf{Limitations.}
Generalized CycleGRPO uses region reconstruction as a verifiable proxy for
caption quality. Reconstructability, however, does not guarantee that a
caption is complete, fluent, or free of every unsupported detail. Our
reference-free validity rules discourage obvious shortcuts, such as leaking
coordinates or timestamps, but cannot enumerate all implicit shortcuts. The
closed loop also assumes that the initial policy can perform both mappings
reasonably well; when its localization is unreliable, the captioning side
receives a noisy learning signal. This dependence helps explain why gains can
vary across initializations and target domains. In addition, the nested
$N\!\times\!K$ rollouts are more expensive than one-directional GRPO. Finally,
DOVE evaluates visually distinguishable events within supplied temporal
intervals. It does not yet cover audio-dependent evidence or require a model
to discover event boundaries.

\noindent\textbf{Future work.}
Future work could complement geometric consistency with
reference-free signals for semantic coverage, factuality, and uncertainty,
while preserving the framework's self-evaluating nature. Adaptive
rollout allocation, trajectory reuse, and distillation could reduce the cost
of the nested cycle as training scales. Beyond the region
spaces studied here, the same principle may extend to audio-visual events,
3D regions, and embodied trajectories. Extending DOVE to these settings and
to joint captioning and boundary discovery would provide a broader test of
whether cycle-trained descriptions remain discriminative in open-world
videos.

%% file: secs/5_conclusion.tex
\section{Conclusion}
\label{sec:conclusion}

Generalized CycleGRPO extends our ECCV CycleGRPO formulation from image masks
to video intervals and spatio-temporal trajectories.
A single policy learns captioning and localization through reciprocal rollouts, using reconstruction quality as supervision without reference captions or a separate critic. 
Across all three settings, the method improves its matched base policies. 
We also introduce DOVE, which tests whether a caption can distinguish one event from repeated
same-query occurrences in a long video. 
Our results show that cycle reconstruction provides a practical, verifiable
training signal whenever a model must both describe and localize visual
content.

%% file: secs/contributors.tex
\par\medskip\noindent
\textbf{Contributors:}
Xin Zhang\textsuperscript{1},
Haochen Wang\textsuperscript{2},
Beibei Lin\textsuperscript{1},
Zhaoxiang Zhang\textsuperscript{2},
Robby T. Tan\textsuperscript{1},
Jason Li\textsuperscript{3}, and
Dacheng Tao\textsuperscript{3},

{\small\noindent
\textsuperscript{1}National University of Singapore;
\textsuperscript{2}University of Chinese Academy of Sciences;
\textsuperscript{3}Nanyang Technological University.\par}

%% file: secs/appendix_overview.tex
\section*{Appendix Overview}
The appendix provides additional implementation details, controlled analyses,
and benchmark statistics that complement the main paper. Specifically,
Appendix~\ref{app:implementation} documents the complete training
configurations for the spatial, temporal, and spatio-temporal settings.
Appendix~\ref{app:spatial-additional} further examines the spatial
instantiation through additional region-captioning and localization examples,
continuous bounding-box outputs, a larger SAMTok-8B backbone, and a
representative closed-loop rollout.
Appendix~\ref{app:temporal-visualization} presents qualitative temporal
grounding comparisons on TimeLens, while
Appendix~\ref{app:vstar-visualization} provides joint temporal--spatial
grounding examples and representative failure cases on V-STaR. Finally,
Appendix~\ref{app:dove-statistics} characterizes DOVE in terms of video and
interval durations, repeated-occurrence structure, query diversity,
temporal-probe coverage, and model-independent difficulty.

%% file: secs/6_appendix_exp_settings.tex
\section{Detailed Experimental Settings}
\label{app:implementation}

This section details the training configurations summarized in
Sec.~\ref{sec:exp}. 
A dash denotes a parameter that is not applicable to the corresponding training stage.
The training code is publicly available.

\noindent\textbf{Temporal training data preparation.}
The source videos and temporal interval annotations come from the OMTG
training data~\cite{xu2026omtg}.
For the Qwen3-VL initialization, we randomly partition source videos into two
equal subsets: one for bidirectional co-SFT and the other for RL. All examples
from a video stay in the same subset, yielding 18,742 RL examples.
For the OMTG initialization, we use an offline VLM-assisted screening procedure
to exclude incomplete or visually ambiguous training intervals from 36,892
candidate examples. Given the annotated target interval, the VLM uses sampled
frames and adjacent context to assess event completeness, with an independently
generated description serving as evidence for screening. It compares the target
with up to two visually nearest intervals sharing the same video and query,
selected using color-histogram and low-frequency image descriptors. Each pair
is evaluated in both presentation orders, using only visual evidence rather
than audio, text overlays, timestamps, or occurrence indices. We retain complete
events only when all selected comparisons consistently identify a visual
difference; singleton groups require a complete event. Ambiguous, incomplete,
uncertain, and failed-screening examples are excluded, leaving 18,735 examples.
This procedure is used only for offline data quality control: the generated
descriptions and judgments are not used as training targets or RL rewards.

\begin{table}[H]
\centering
\scriptsize
\caption{Implementation details for the spatial setting. The configuration is
carried over unchanged from the ECCV version~\cite{zhang2026cyclegrpo}.}
\label{tab:app-spatial-settings}
\setlength{\tabcolsep}{6pt}
\renewcommand{\arraystretch}{1.0}
\begin{tabular}{p{0.24\textwidth}p{0.68\textwidth}}
\toprule
Setting & Configuration \\
\midrule
Initialization & Qwen3-VL-4B-SAMTok~\cite{zhou2026samtok} \\
Training data & 20K DenseWorld~\cite{li2025denseworld} images with object masks and 1K GRES~\cite{gres} no-target expressions \\
Region representation & Discrete mask tokens in both prompts and responses \\
Input prompt & \texttt{<image>} ``Provide a detailed description of this region \texttt{<mask tokens>}.'' When a zoomed view is available, the prompt continues: ``Zoom in with the perspective as \texttt{<image>}, \texttt{<mask tokens>}.'' \\
Training objective & Spatial mask--text--mask CycleGRPO \\
Trainable modules & Language model and multimodal projector; visual encoder frozen \\
Precision and memory & bfloat16 with gradient checkpointing \\
Training length & 1 epoch \\
Optimizer & AdamW \\
Learning rate / weight decay & $1\times10^{-6}$ / $1\times10^{-2}$ \\
Global batch size & 128 \\
Rollouts & $N=6$ caption samples and $K=6$ localization rollouts per caption \\
Distributed training & FSDP with the extended \texttt{verl} codebase \\
\bottomrule
\end{tabular}
\end{table}

\begin{table}[H]
\centering
\scriptsize
\caption{Training configurations for temporal CycleGRPO with
Qwen3-VL~\cite{Qwen3-VL} and OMTG~\cite{xu2026omtg} initializations.}
\label{tab:app-temporal-settings}
\setlength{\tabcolsep}{6pt}
\renewcommand{\arraystretch}{1.0}
\begin{tabular}{p{0.22\textwidth}p{0.34\textwidth}p{0.34\textwidth}}
\toprule
Setting & Qwen3-VL initialization & OMTG initialization \\
\midrule
Initialization & Qwen3-VL-4B~\cite{Qwen3-VL} with bidirectional co-SFT & OMTG-4B RL~\cite{xu2026omtg} with bidirectional co-SFT \\
RL training examples & 18,742 & 18,735 \\
Training data & OMTG~\cite{xu2026omtg}: video-disjoint RL split comprising the remaining 50\% of source videos & OMTG~\cite{xu2026omtg}: filtered visually distinct temporal examples \\
Rollout / actor update batch size & 128 / 128 & 64 / 64 \\
Caption rollouts $N$ & 4 & 4 \\
Localization rollouts $K$ & 4 & 4 \\
Video sampling & 0.25 fps & 2 fps \\
Maximum pixels/frame & 262,144 & 786,432 \\
Prompt / response limits & 8,192 / 8,192 tokens & 16,384 / 768 tokens \\
\midrule
Objective & \multicolumn{2}{p{0.68\textwidth}}{Interval-to-caption-to-interval cycle} \\
Reward / supervision & \multicolumn{2}{p{0.68\textwidth}}{Reconstruction tIoU with output-validity constraints; no reference caption or external language-model reward judge} \\
Captioning input & \multicolumn{2}{p{0.68\textwidth}}{Source video and a prompt specifying the target time interval} \\
Optimizer & \multicolumn{2}{p{0.68\textwidth}}{AdamW} \\
Learning rate / weight decay & \multicolumn{2}{p{0.68\textwidth}}{$1\times10^{-6}$ / $1\times10^{-2}$} \\
KL coefficient & \multicolumn{2}{p{0.68\textwidth}}{$1\times10^{-2}$} \\
Policy ratio clipping & \multicolumn{2}{p{0.68\textwidth}}{$[0.8,1.3]$; dual-clipping coefficient $c=3$} \\
Loss aggregation & \multicolumn{2}{p{0.68\textwidth}}{Mean over valid response tokens} \\
Trainable modules & \multicolumn{2}{p{0.68\textwidth}}{Language model and multimodal projector; visual encoder frozen} \\
Precision / memory & \multicolumn{2}{p{0.68\textwidth}}{bfloat16 / gradient checkpointing} \\
Distributed training & \multicolumn{2}{p{0.68\textwidth}}{FSDP with the extended \texttt{verl} codebase} \\
\bottomrule
\end{tabular}
\end{table}

\begin{table}[H]
\centering
\scriptsize
\caption{Implementation details for the Open-o3-Video-4B mixed-task training
used in the main spatio-temporal and general-video results.}
\label{tab:app-st-settings}
\setlength{\tabcolsep}{6pt}
\renewcommand{\arraystretch}{1.0}
\begin{tabular}{p{0.24\textwidth}p{0.68\textwidth}}
\toprule
Setting & Configuration \\
\midrule
Initialization & Open-o3-Video-4B (RL)~\cite{meng2025openo3video}, followed by one epoch of inverse-caption and video-QA co-SFT \\
Training data & Open-o3-Video~\cite{meng2025openo3video}: 35,887 examples comprising 4,143 spatial, 4,820 temporal, 11,924 spatio-temporal, and 15,000 general video-QA examples \\
Region representations & Normalized image boxes, temporal intervals, and sparse timestamp--box supports \\
Training objectives & Region--text--region CycleGRPO for the three region tasks; answer-based rewards for auxiliary video QA \\
Trainable modules & Language model and multimodal projector; visual encoder frozen \\
Precision / memory & bfloat16 / gradient checkpointing \\
Optimizer / schedule & AdamW; cosine learning-rate schedule \\
Learning rate / weight decay & $1\times10^{-6}$ / $1\times10^{-2}$ \\
Rollout / actor update batch size & 128 / 128 \\
Caption rollouts $N$ & 4 \\
Localization rollouts $K$ & 4 \\
KL coefficient & $1\times10^{-2}$ \\
Policy ratio clipping & $[0.8,1.3]$; dual-clipping coefficient $c=3$ \\
Loss aggregation & Mean over valid response tokens \\
Sampling temperature / top-$p$ & 1.0 / 1.0 \\
Prompt / response limit & 16,384 / 768 tokens \\
Video input & 2 fps configuration; spatio-temporal examples specify 16 frames; maximum 262,144 pixels per frame \\
Spatio-temporal caption prompt & \texttt{<video>} followed by the target timestamp--box support and: ``Describe only the visible target and event precisely enough to relocate them. Do not mention object tags, coordinates, timestamps, seconds, or frame indices. Output description only.'' \\
Temporal reward shaping & Additional temporal-proximity score with coefficient 2.0, computed from target and reconstructed temporal supports \\
Language supervision & No reference captions for the cycle branches; reference answers for auxiliary video QA; no external language-model reward judge \\
Distributed training & FSDP with the extended \texttt{verl} codebase \\
\bottomrule
\end{tabular}
\end{table}

In the Open-o3-Video mixed-task setting (Tab.~\ref{tab:app-st-settings}),
auxiliary video-QA multiple-choice answers
receive a correctness reward and free-form answers receive a reference-based
ROUGE reward, together with format and repetition checks. Temporal reward
shaping uses interval-center, boundary, and duration agreement to supplement
overlap-based feedback.

\FloatBarrier
\section{Additional Spatial Analyses}
\label{app:spatial-additional}

\begin{figure}[H]
    \centering
    \includegraphics[width=0.96\textwidth]{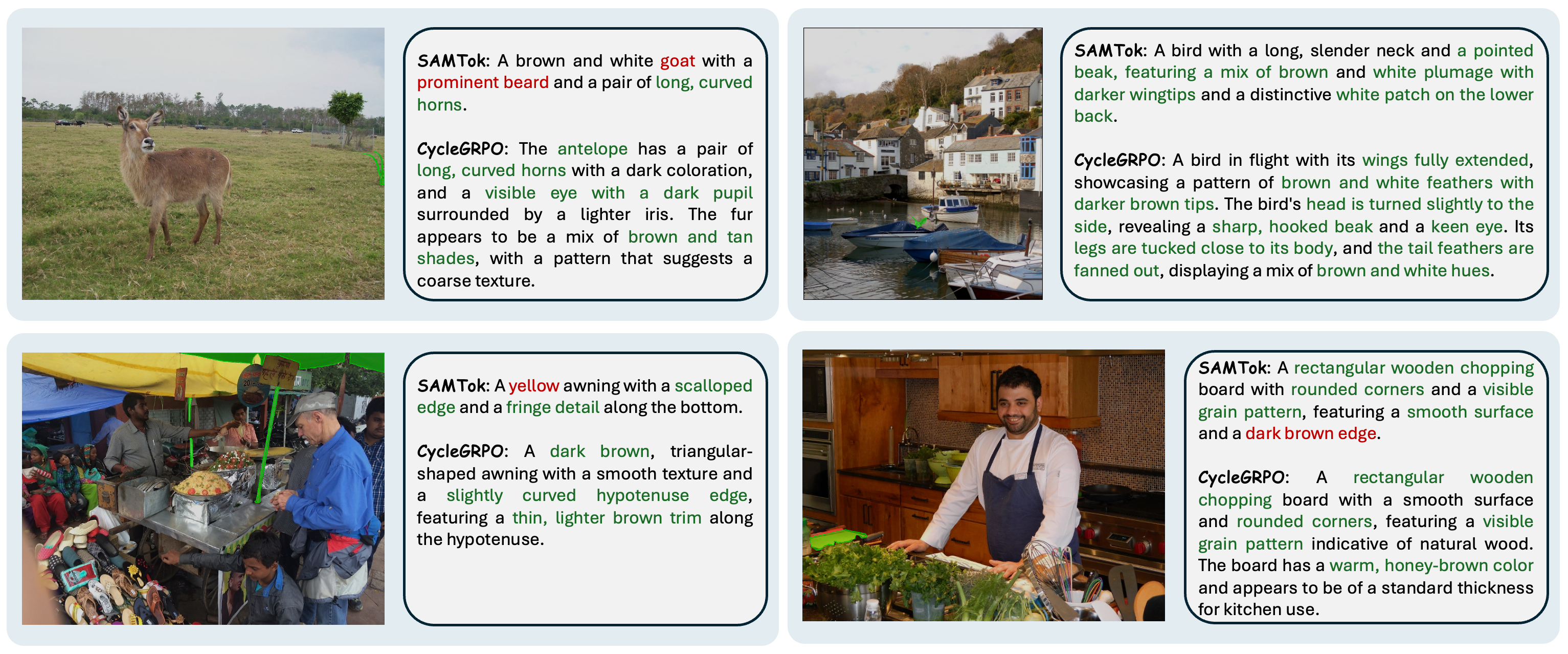}
    \caption{Qualitative comparison of region captioning on
    DLC-Bench~\cite{dam} between SAMTok~\cite{zhou2026samtok} and CycleGRPO.
    Target regions are highlighted in green. Within the captions, green text
    marks accurate attributes and red text marks errors.}
    \label{fig:app-eccv-captioning}
\end{figure}

\begin{figure}[H]
    \centering
    \includegraphics[width=0.96\textwidth]{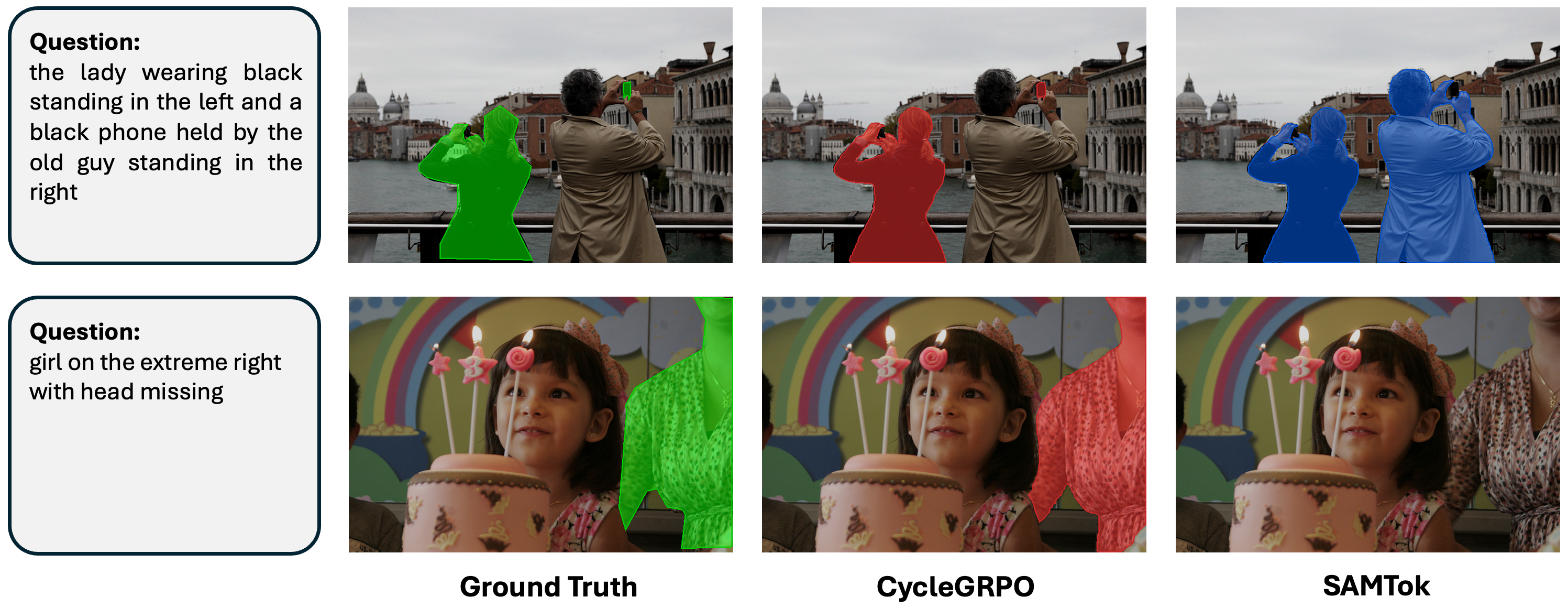}
    \caption{Qualitative mask-localization results on GRES~\cite{gres}.
    CycleGRPO is compared with SAMTok~\cite{zhou2026samtok} and the ground
    truth.}
    \label{fig:app-eccv-gres}
\end{figure}

\begin{figure}[H]
    \centering
    \includegraphics[width=0.96\textwidth]{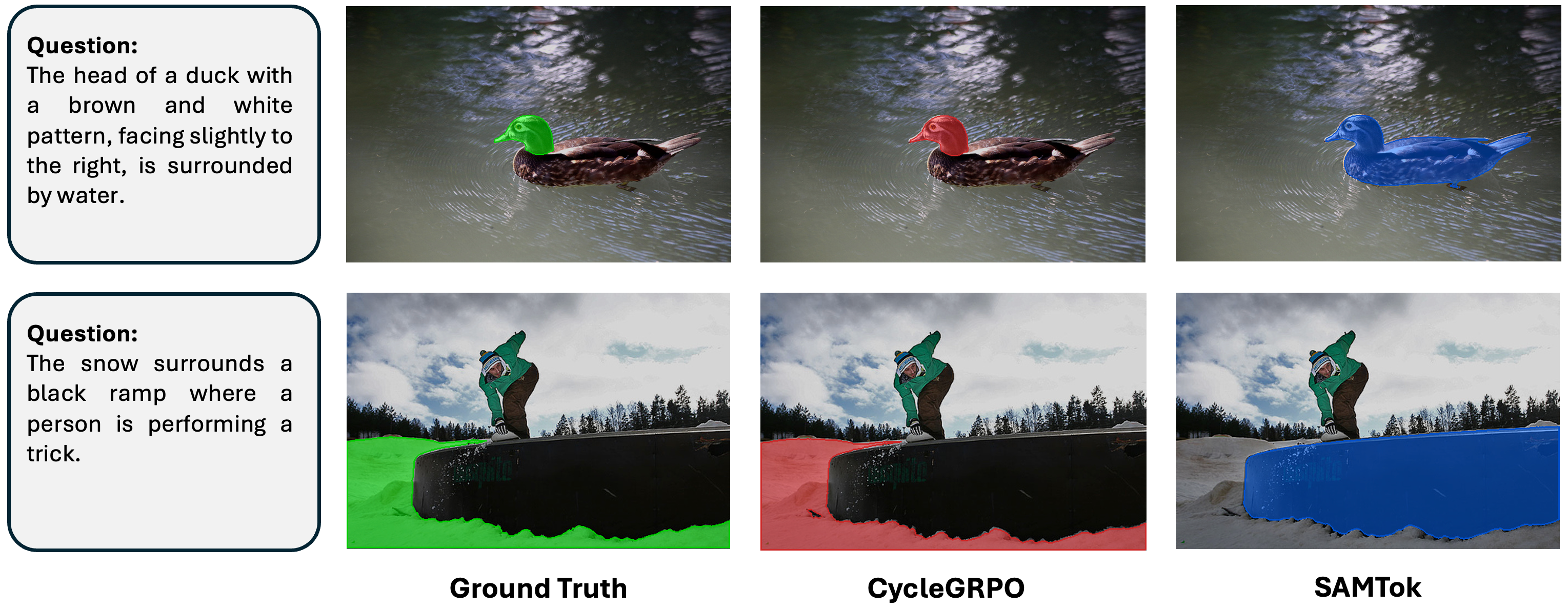}
    \caption{Qualitative mask-localization results on
    GroundingSuite~\cite{zhou2026samtok}. The first row evaluates part-level
    segmentation; the second evaluates amorphous stuff regions. Predictions
    from SAMTok and CycleGRPO are compared with the ground truth.}
    \label{fig:app-eccv-groundingsuite}
\end{figure}

\begin{figure}[H]
    \centering
    \includegraphics[width=0.70\textwidth]{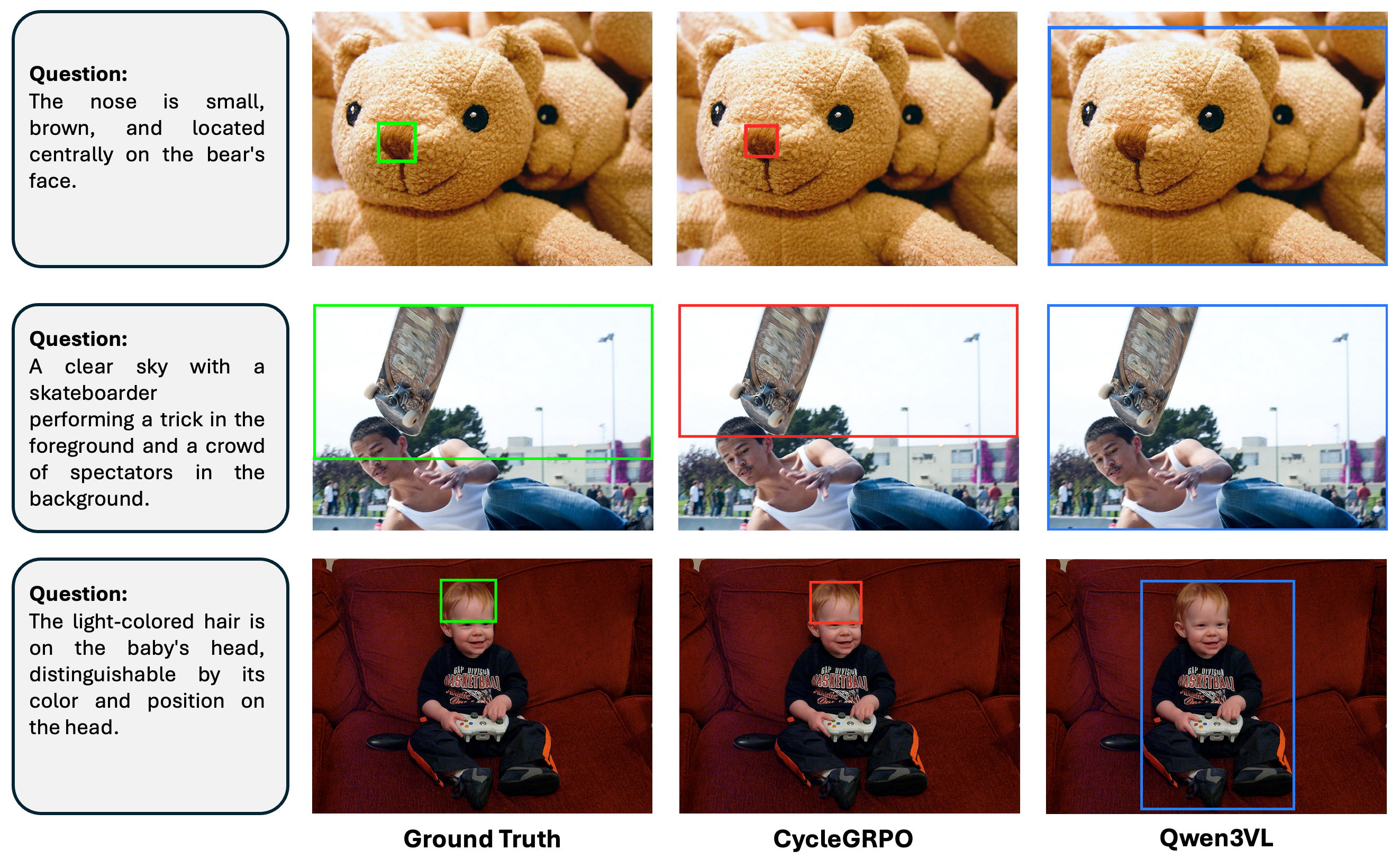}
    \caption{Qualitative GroundingSuite results with continuous bounding-box
    outputs. CycleGRPO predictions, Qwen3-VL predictions, and ground-truth
    boxes are shown in red, blue, and green, respectively.}
    \label{fig:app-eccv-groundingsuite-bbox}
\end{figure}

\noindent
\textbf{Additional qualitative comparisons.}
Figs.~\ref{fig:app-eccv-captioning}--\ref{fig:app-eccv-groundingsuite-bbox}
provide the complete spatial qualitative comparisons from our ECCV version.
They complement the cross-domain examples in Fig.~\ref{fig:visual-analysis}
with region captioning, mask localization, and continuous bounding-box
localization cases.

In Fig.~\ref{fig:app-eccv-captioning}, CycleGRPO captures the target's texture,
shape, and color more accurately than SAMTok. The GRES examples in
Fig.~\ref{fig:app-eccv-gres} show localization of disjoint entities without
including the man's coat, and recovery of an occluded target missed by the
baseline. Fig.~\ref{fig:app-eccv-groundingsuite} covers both part-level targets,
such as the duck's head, and amorphous stuff regions, such as snow.
For continuous boxes in Fig.~\ref{fig:app-eccv-groundingsuite-bbox}, CycleGRPO
localizes the bear's nose, the baby's hair, and the clear sky more tightly;
the baseline boxes include more of the surrounding subject.

\begin{figure}[H]
    \centering
    \includegraphics[width=\textwidth,trim=31bp 7bp 17bp 7bp,clip]{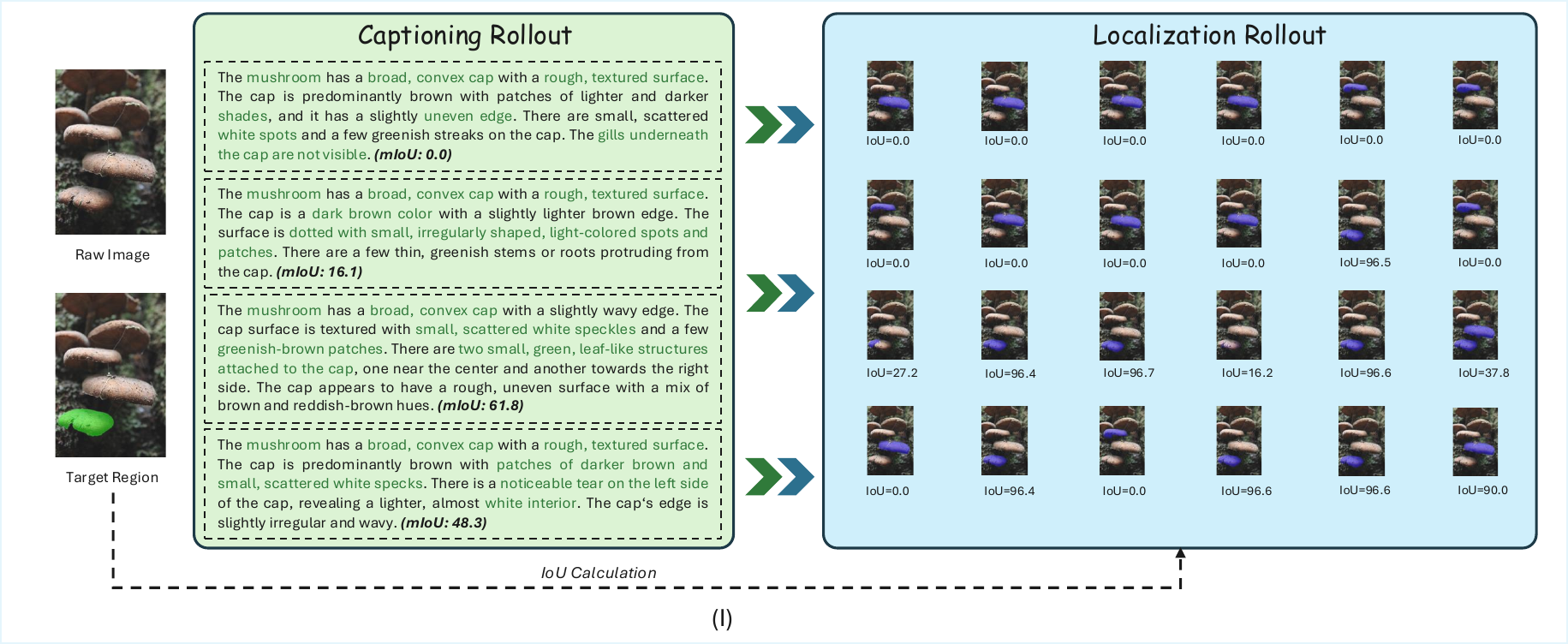}
    \caption{A spatial region--text--region rollout. Each of four descriptions of the
    target mushroom conditions six localization rollouts. Green and red
    text mark accurate and erroneous evidence; blue masks show the reconstructed
    regions. More precise descriptions generally yield higher reconstruction
    IoU.}
    \label{fig:app-spatial-rollout}
\end{figure}

\noindent
\textbf{Scaling the SAMTok backbone.}
We additionally apply CycleGRPO to SAMTok initialized from Qwen3-VL-8B.
As shown in Tab.~\ref{tab:app-spatial-samtok8b}, the method improves the
DLC-Bench average from 63.7\% to 68.1\% and the GroundingSuite average gIoU
from 63.1\% to 64.5\%, confirming that the gains persist with a larger spatial
backbone.

\begin{center}
\small
\captionof{table}{CycleGRPO with SAMTok initialized from Qwen3-VL-8B, evaluated on
DLC-Bench~\cite{dam} and GroundingSuite~\cite{zhou2026samtok}. All values are
percentages.}
\label{tab:app-spatial-samtok8b}
\setlength{\tabcolsep}{3pt}
\resizebox{0.65\textwidth}{!}{%
\begin{tabular}{lccc|ccccc}
\toprule
\multirow{2}{*}{Method} & \multicolumn{3}{c|}{DLC-Bench} & \multicolumn{5}{c}{GroundingSuite (gIoU)} \\
\cmidrule(lr){2-4}\cmidrule(lr){5-9}
& Pos. & Neg. & Avg. & Single & Stuff & Multi & Part & Avg. \\
\midrule
SAMTok-8B & 47.3 & 80.0 & 63.7 & 88.6 & \textbf{15.9} & 65.9 & \textbf{58.7} & 63.1 \\
\rowcolor{blue!8}\textbf{\quad + Ours} & \textbf{51.6} & \textbf{84.6} & \textbf{68.1} & \textbf{90.1} & 15.2 & \textbf{77.7} & 55.2 & \textbf{64.5} \\
\bottomrule
\end{tabular}}
\end{center}

\noindent
\textbf{Closed-loop rollout.}
Fig.~\ref{fig:app-spatial-rollout} provides a representative training case.
Descriptions that preserve distinctive visual evidence reconstruct the target
region reliably, whereas hallucinated or insufficiently discriminative details
produce low-overlap localization rollouts. This correspondence turns the
model's own reconstruction accuracy into a direct learning signal for region
description quality.

\FloatBarrier

\section{Temporal Grounding Visualization}
\label{app:temporal-visualization}

\begin{figure}[H]
    \centering
    \includegraphics[width=0.90\textwidth]{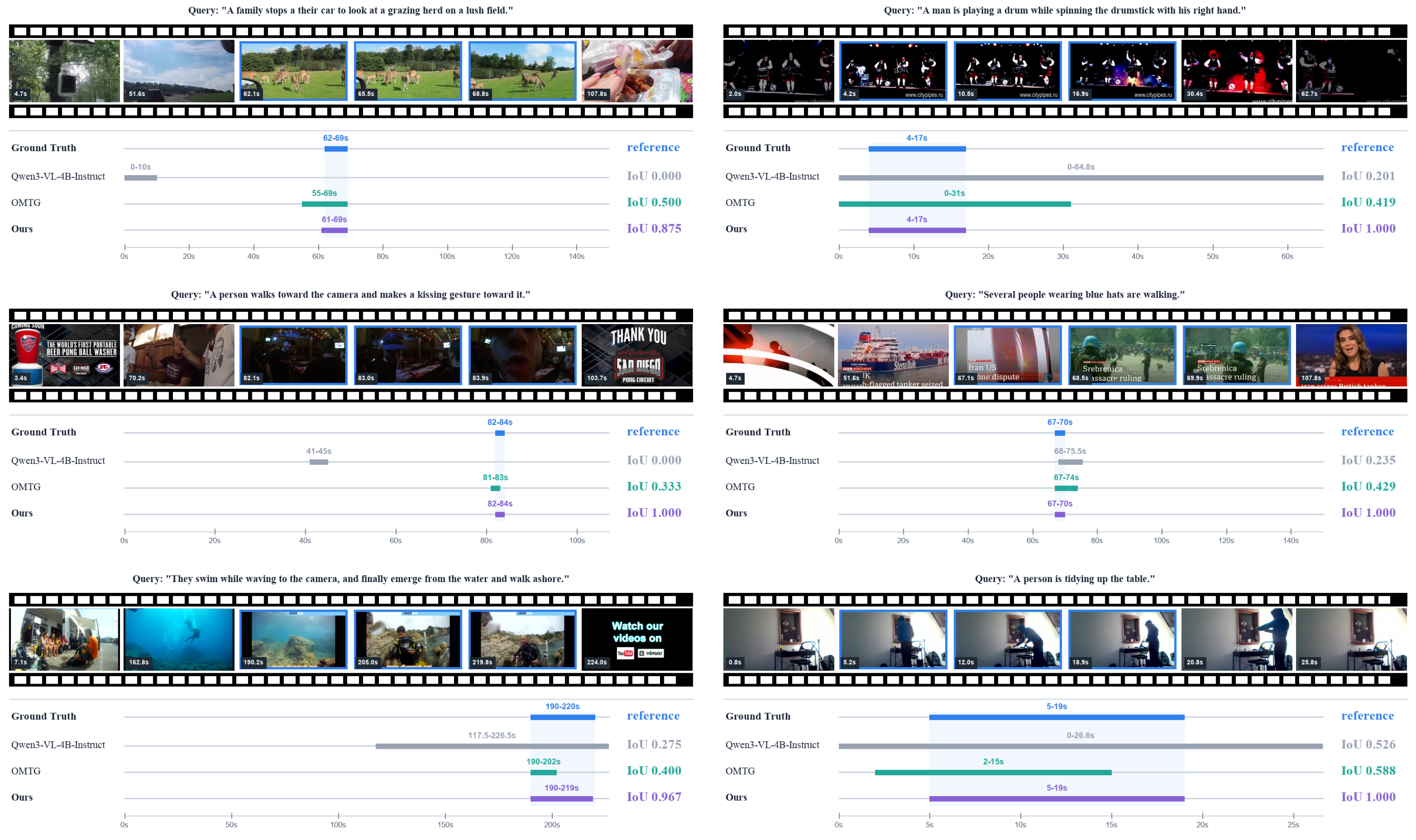}
    \caption{Qualitative temporal grounding comparisons on
    TimeLens~\cite{zhang2026timelens}. Blue-bordered frames depict the reference event,
    and the timelines compare the reference interval with predictions from
    Qwen3-VL-4B-Instruct, OMTG, and our model. The displayed IoU values denote
    temporal IoU.}
    \label{fig:app-timelens-visualization}
\end{figure}

\noindent
\textbf{Boundary-sensitive temporal localization.}
Fig.~\ref{fig:app-timelens-visualization} covers events with substantially
different durations, from brief gestures and object interactions to longer
activities. 
The base model frequently misses the target or predicts an overly broad interval, while OMTG generally narrows the prediction but still includes irrelevant context or truncates part of the event. 
In these examples, our model aligns the predicted boundaries more closely with the reference, including short events
embedded in long videos and activities whose completion must be distinguished
from their preceding context.

\section{Additional V-STaR Qualitative Analysis}
\label{app:vstar-visualization}

\begin{figure}[H]
    \centering
    \includegraphics[width=0.94\textwidth]{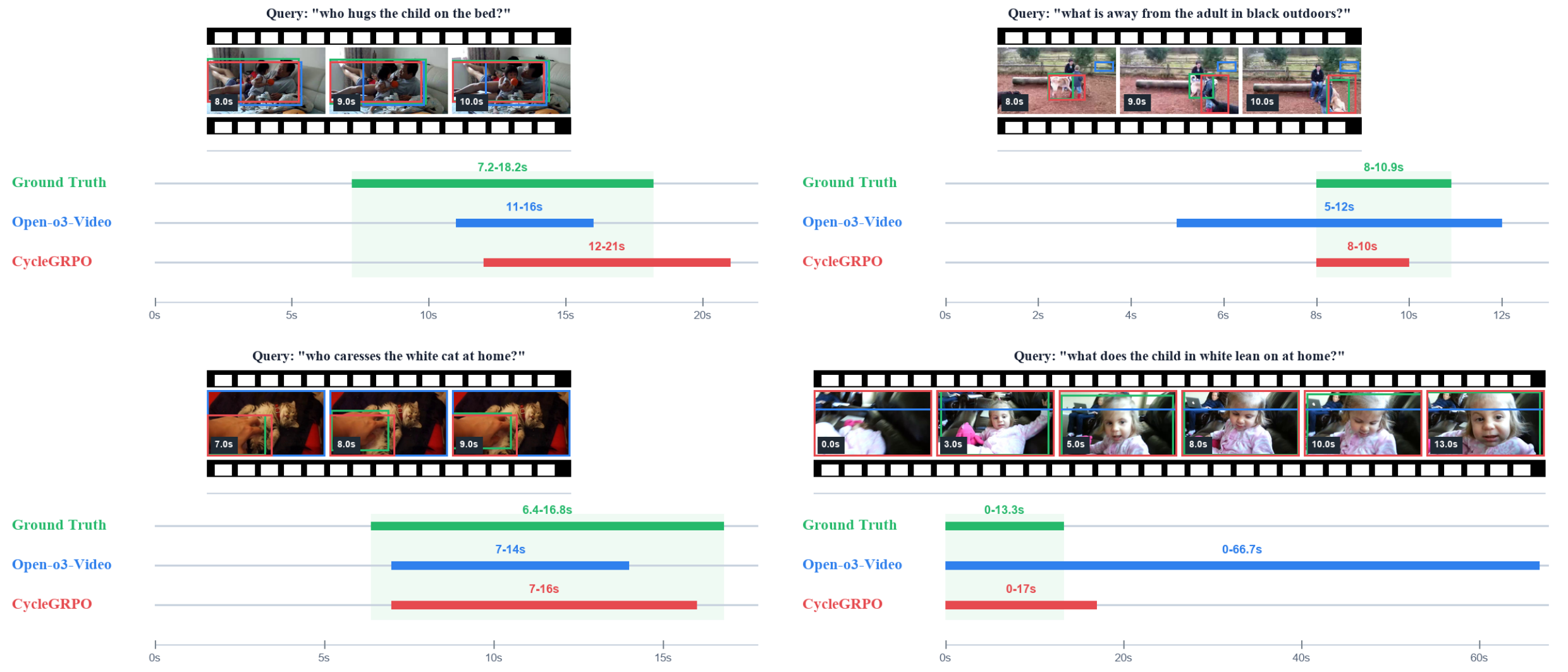}
    \caption{Qualitative V-STaR~\cite{cheng2025vstar} examples for joint
    temporal--spatial grounding. Each panel contains a query, sampled frames
    with the associated object evidence, and the predicted temporal interval.
    The green bar and shaded band denote the reference interval; blue and red
    bars denote Open-o3-Video and Generalized CycleGRPO, respectively.}
    \label{fig:app-vstar-visualization}
\end{figure}

\noindent
\textbf{Joint temporal--spatial evidence.}
Fig.~\ref{fig:app-vstar-visualization} makes the temporal component of
V-STaR's joint evidence requirement explicit. For the outdoor query in the
upper-right panel, Open-o3-Video spans 5--12\,s for a reference event from
8--10.9\,s, whereas CycleGRPO restricts the prediction to 8--10\,s. Likewise,
for the child-in-white example, the baseline extends to 66.7\,s after the
reference event ends at 13.3\,s; our prediction remains close to the event
and stops at 17\,s. The lower-left example further shows a more complete
recovery of the event's end boundary. These examples align with the aggregate
V-STaR gains in Tab.~\ref{tab:main_vstar}: learning to reconstruct the
queried temporal--spatial region discourages explanations supported by the
right object at the wrong time or by irrelevant temporal context. At the same
time, the upper-left example remains difficult: the predicted interval is
late and overextended, showing that cycle consistency improves but does not
guarantee boundary accuracy in every case.

\begin{figure}[H]
    \centering
    \includegraphics[width=0.94\textwidth]{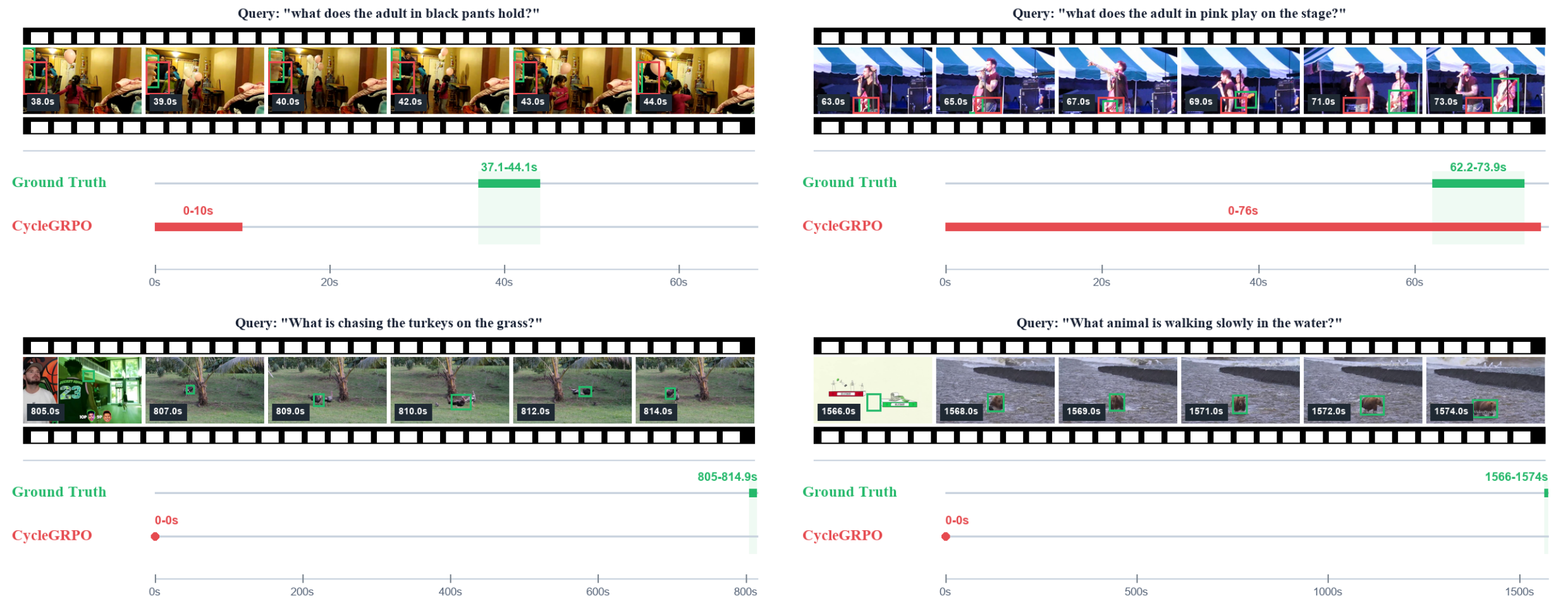}
    \caption{Representative failure cases of Generalized CycleGRPO on
    V-STaR~\cite{cheng2025vstar}. From upper left to lower right, the examples
    show incorrect temporal retrieval despite a correct semantic answer,
    imprecise spatial localization with an over-broad temporal prediction,
    semantic hallucination for an unusual interaction, and a missed event near
    the end of a long video. Green and red denote the ground truth and
    CycleGRPO predictions, respectively.}
    \label{fig:app-vstar-failures}
\end{figure}

\noindent
\textbf{Representative failure modes.}
Fig.~\ref{fig:app-vstar-failures} exposes limitations that are obscured by
aggregate scores. In the upper-left case, the model correctly identifies the
cup but places the event at 0--10\,s rather than 37.1--44.1\,s. In the
upper-right case, it recognizes the red electric guitar, yet predicts nearly
the entire video and fails to localize the object precisely. The lower-left
case confuses a sea otter chasing two turkeys with a man chasing a deer,
illustrating semantic hallucination under an unusual interaction. Finally, the
lower-right case misses a bear appearing near the end of a long video and
incorrectly concludes that no animal is present. These examples highlight
remaining challenges in semantic robustness, fine-grained spatial grounding,
and long-range temporal retrieval.

\section{DOVE Statistics}
\label{app:dove-statistics}

\begin{figure}[H]
    \centering
    \includegraphics[width=0.90\textwidth]{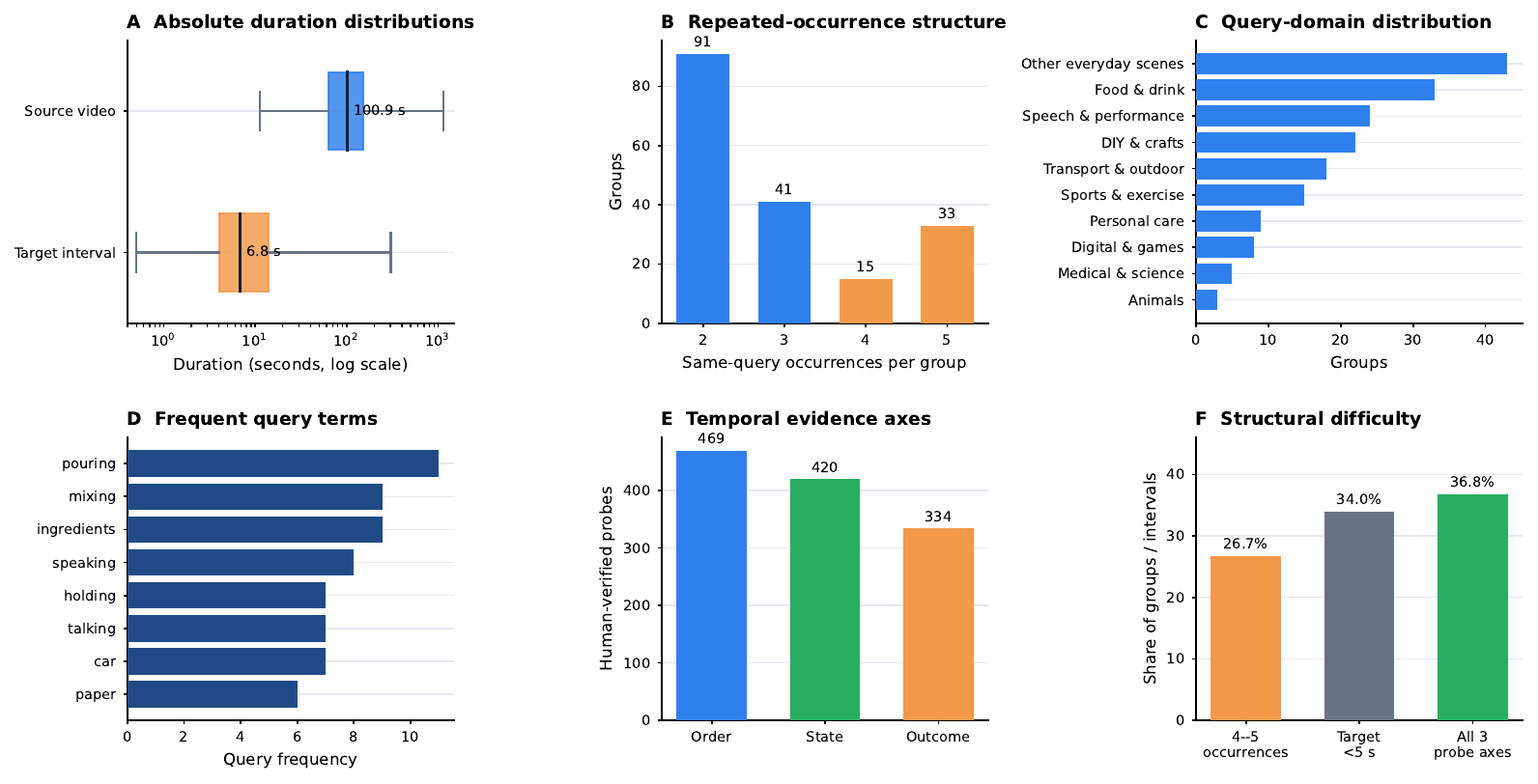}
    \caption{Statistics of DOVE. (A) Absolute source-video
    and target-interval duration distributions on a logarithmic time axis (seconds);
    boxes show the interquartile ranges and whiskers show the full ranges. (B) Number
    of same-query occurrences per group. (C) Query-domain distribution from a
    deterministic lexical taxonomy over the human-reviewed queries. (D) Most
    frequent non-actor query terms. (E) Human-verified temporal probes by
    evidence axis. (F) Model-independent difficulty indicators: groups with
    four or five occurrences, target intervals shorter than five seconds, and
    intervals covered by all three probe axes.}
    \label{fig:dove-statistics}
\end{figure}

\noindent
\textbf{Diversity and structural difficulty.}
Fig.~\ref{fig:dove-statistics} characterizes the benchmark beyond its scale.
The median source video lasts 100.9 seconds, whereas the median target interval
lasts 6.8 seconds; Fig.~\ref{fig:dove-statistics}(A) shows both duration distributions on a logarithmic axis.
The queries span food and drink, speech and performance, crafts, transportation and outdoor activity, sports, personal care, digital interfaces, medical scenes, and animals; the
plotted categories come from a reproducible keyword taxonomy rather than
additional human labels. 
Structurally, 48 groups (26.7\%) contain four or five same-query occurrences, and 180 targets (34.0\%) last less than five seconds.
The 1,223 probes provide a median of two probes per covered interval, while 195
intervals (36.8\%) have probes covering all three types: Order, State, and Outcome.
Thus, DOVE combines short targets, repeated same-query distractors, and
multi-axis temporal evidence without defining difficulty from any evaluated
model's outputs.